\documentclass[11pt]{article}

\usepackage[final]{acl}

\usepackage{times}
\usepackage{latexsym}
\usepackage{tabularx}
\usepackage[T1]{fontenc}
\usepackage[utf8]{inputenc}

\usepackage{microtype}

\usepackage{inconsolata}

\usepackage{graphicx}
\usepackage{algorithmicx}
\usepackage{algpseudocode}
\newcommand{\toolname}{\textsc{SkillPrivus}}
\usepackage[utf8]{inputenc} % allow utf-8 input
\usepackage[T1]{fontenc}    % use 8-bit T1 fonts
\usepackage{hyperref}       % hyperlinks
\usepackage{url}            % simple URL typesetting
\usepackage{booktabs}       % professional-quality tables
\usepackage{amsfonts}       % blackboard math symbols
\usepackage{amssymb}        % \varnothing and other symbols
\usepackage{nicefrac}       % compact symbols for 1/2, etc.
\usepackage{microtype}      % microtypography
\usepackage{xcolor}         % colors
\usepackage[most]{tcolorbox} % highlighted qualitative-example and prompt boxes
\tcbuselibrary{listings}    % breakable verbatim listings inside tcolorbox (appendix prompt boxes)
\usepackage{fvextra}        % extends fancyvrb: Verbatim with breaklines/breakanywhere
\usepackage{amsmath}        % math environments
\usepackage{algorithm}      % floating algorithm environment
\usepackage{algpseudocode}  % pseudocode (algorithmicx)
\usepackage{tikz}           % vector framework figure (no image conversion needed)
\usetikzlibrary{arrows.meta,positioning,fit,backgrounds,calc,shapes.geometric}
\tcbuselibrary{listings,skins,breakable}
\usepackage{booktabs}

\usepackage{pgfplots}  
\pgfplotsset{compat=1.18}
\title{SkillPrivus: Uncovering \underline{PRIV}acy of \underline{US}e Harms in Agentic Skills}

\author{
  Mariana Fernandez-Espinosa$^{1*}$,
  Ivoline Ngong$^{2*}$, \\
  \textbf{Keerthiram Murugesan$^{2}$,
  Karthikeyan Natesan Ramamurthy$^{2}$} \\[0.5em]
  $^{1}$University of Notre Dame \\
  $^{2}$IBM Research \\
  \texttt{mferna23@nd.edu} \\
  \texttt{\{ivoline,keerthiram.murugesan\}@ibm.com} \\
  \texttt{knatesa@us.ibm.com} \\[0.3em]
  {\small $^{*}$Equal contribution.}
}

\usepackage{tikz}

\definecolor{skillnavy}{HTML}{253B53}
\definecolor{skillorange}{HTML}{F8D9AF}

\begin{document}
 \maketitle
\begin{abstract}
Agent skills extend AI assistants through installable instructions and code that can access users’ files, accounts, and data. Even when performing their advertised tasks, skills may infer sensitive attributes, retain unnecessary information, or disclose data to unexpected recipients. We study these potential privacy-of-use harms through \textbf{\toolname{}}, a static audit framework that analyzes skills before installation without executing them. \toolname{} extracts data operations from code and natural-language instructions, grounds them in source evidence, and adapts Contextual Integrity to assess their appropriateness. Applied to 726 real skills from public marketplaces, 563 of which handle personal data, \toolname{} flags at least one potential violation in 307 of these skills (54.5\%), and its judgments agree with three blind human annotators on 83.2\% ($\kappa=0.35$) of resolvable moves. The flagged behaviors include sensitive inferences, information use beyond the stated task, and disclosures to unexpected recipients, suggesting that privacy harms can arise from ordinary skill functionality.
\end{abstract}

\section{Introduction}

\begin{figure}[t!]
  \centering
  \includegraphics[width=0.90\columnwidth]{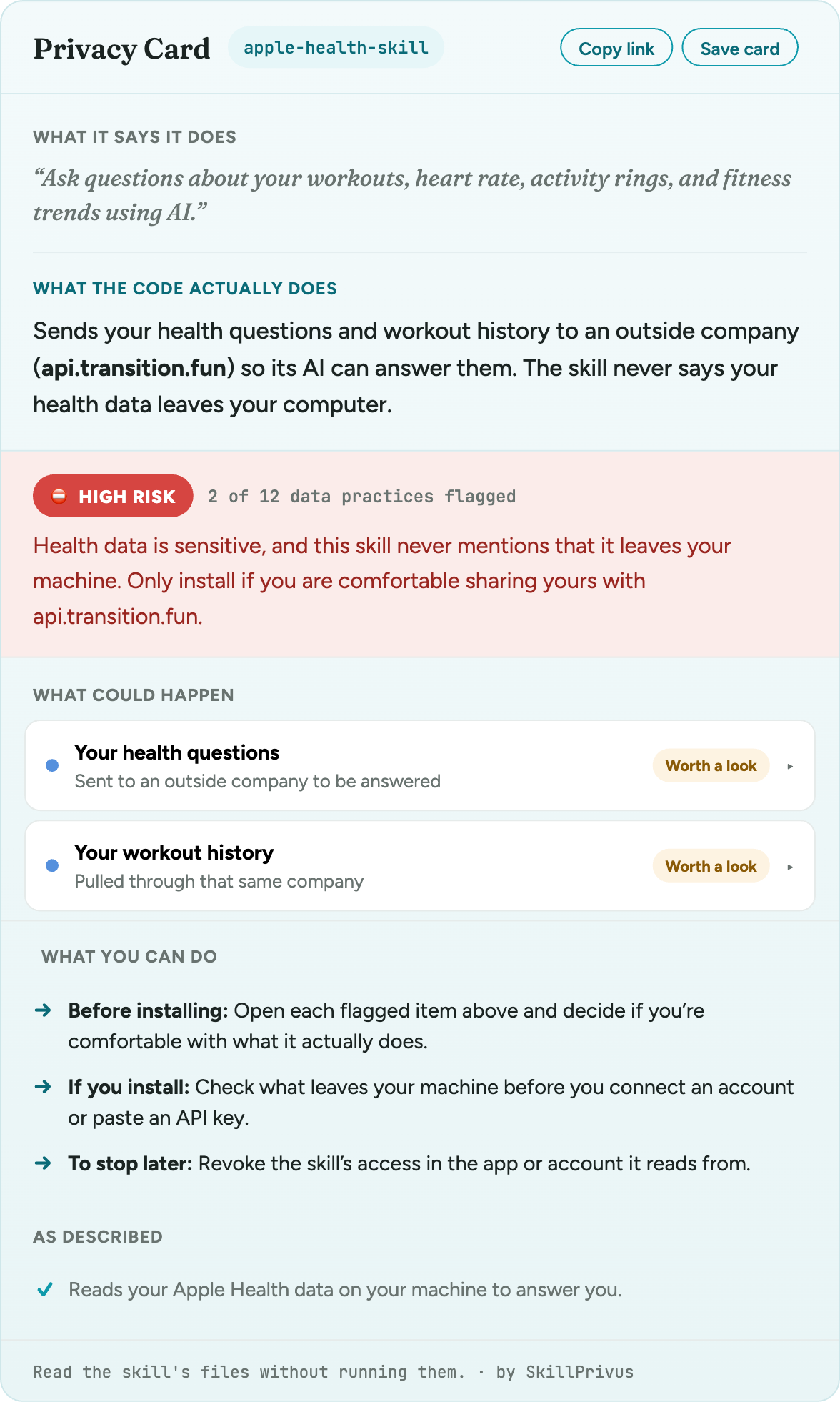}
  \caption{The \toolname{} \emph{Privacy Card}, the output of a static
  audit, showing (\textsc{what it says it does}) vs (\textsc{what the code
  actually does}) for a real \texttt{apple-health} skill collected from a
  public marketplace.}
  \label{fig:privacycard}
\end{figure}

% AI assistants increasingly extend their capabilities through ...
% \begin{figure}[t]
%     \centering
%     \includegraphics[width=\columnwidth]{figures/mainfigure.pdf}
%     \caption{Overview of \toolname{}. The pipeline
%     (1) screens skills for personal-data handling,
%     (2) extracts evidence-linked data moves, and
%     (3) judges their appropriateness against the stated
%     purpose and Contextual Integrity criteria.
%     The report links potential harms to source evidence,
%     without executing the skill.}
%     \label{fig:pipeline}
% \end{figure}

AI assistants increasingly extend their capabilities through \emph{skills}, installable packages of reusable instructions and code for specialized tasks \citep{liu2026well}. Skills operate through the assistant's available permissions to process users' files, accounts, and personal data. Users decide whether to install a skill based on a short description, which may not fully disclose how it handles these data. A skill can fulfill its advertised purpose while inferring sensitive attributes, retaining unnecessary information, or disclosing data to unexpected recipients. For instance, a skill promising to let users ``talk to their Apple Health data'' ... includes instructions to send health-related questions to a third-party service not identified in its description (Figure~\ref{fig:privacycard}). Such behavior need not involve an attacker or prevent task completion, yet may exceed the data handling that users reasonably expect.

Prior work on skill-related risks has largely adopted a security perspective, examining threats such as prompt injection, tool poisoning, and data exfiltration by compromised tools~\citep{huang2026model}. We focus on \textbf{honest-but-negligent} behavior, where a skill fulfills its stated task but infers, retains, or transmits data beyond what that task warrants, without requiring the adversarial intent of the classical honest-but-curious model. We call these potential harms \emph{privacy of use}, motivating contextual privacy assessment alongside security analysis.

Because these harms can accompany successful task completion, they may remain unnoticed by users. Contextual Integrity provides a foundation for such an assessment by treating privacy as the appropriateness of information flows relative to contextual norms \citep{nissenbaum2004privacy,barth2006privacy}. Under this view, an operation cannot be judged solely from the presence of personal information or an external destination. For agent skills, the stated purpose provides an additional reference for examining whether the implementation exceeds the functionality communicated to the user. This motivates our research question:
\textit{ Can privacy harms in agent skills be detected \emph{before installation} from their artifacts alone, and how prevalent are these harms in non-malicious, functional skills?}

To tackle this, we design \toolname, a static privacy audit framework that analyzes skills without executing them. It screens for personal-data handling, decomposes code and instructions into source-grounded \emph{data moves}, and adapts Contextual Integrity to assess each move against the skill's stated purpose and contextual norms. Potential privacy harms are classified using Solove's taxonomy~\citep{solove2006taxonomy}, which we extend with \emph{inference} to capture newly derived sensitive attributes.

We apply \toolname{} to 726 skills in the wild collected from popular public marketplaces and official repositories, of which 563 handle privacy-relevant data, \toolname{} flags at least one potential violation in 307 of them, which is 54.5\% of the privacy-relevant skills. These are ordinary skills that a security scanner would pass as clean, since none of the violations involve an attacker, and the most common one is simply data reaching a recipient the skill's stated purpose does not support. On a blind human-annotated reference set, \toolname's verdicts agree with three human annotators on 83.2\% of moves ($\kappa=0.346$), substantial agreement for a judgment that human experts themselves contest. Our contributions are: \textbf{(1) Privacy of use.} We are, to our knowledge, the first to study agent skills through privacy rather than security, framing their harms as honest-but-negligent behavior that no security audit is built to catch. \textbf{(2) \toolname{},} a static audit pipeline applying Contextual Integrity and Solove's taxonomy to code and natural-language artifacts, with findings grounded in source evidence. \textbf{(3) A first prevalence measurement} over 563 privacy-relevant public skills, validated against blind human annotations.

\section{Related Work}
\paragraph{Security of the agent-skill ecosystem.}
The growing adoption of agent skills has motivated research on the security of their instructions and supporting code. Large-scale empirical studies document vulnerabilities and malicious behavior across tens of thousands of public skills~\citep{liu2026agents}, and related work analyzes how skill implementations expose credentials~\citep{chen2026credentials} and the threats arising throughout the skill lifecycle~\citep{li2026towards}. A parallel strand treats the problem as \emph{over-privilege}; skills request or exercise more access than they need, and the fix is least-privilege enforcement~\citep{wu2026skillscope} or a benchmark for how minimally a skill operates~\citep{li2026fortis}. That entire lens asks whether a component is malicious or over-scoped, which is orthogonal to our question. A skill can pass every such check, with no injection, no malware, and only the access it was granted, and still harm privacy by using that access for more than the user was told; with no adversary to find and no permission to revoke, the honest-but-negligent skills we study fall outside its threat model by construction.

\paragraph{Auditing description versus behavior.}
Comparing whether software does what its description claims has a substantial history in mobile application analysis. Prior work examines whether implemented behavior and requested permissions align with advertised functionality~\citep{gorla2014chabada, qu2014autocog}. This consistency perspective also extends to privacy disclosures by comparing data flows with policy statements while accounting for the entities receiving the data~\citep{andow2020policheck}.

For agent skills, specifications contain operational instructions while also shaping users' expectations about functionality. Research on skill comprehension finds that these specifications often provide limited information about capability boundaries and expected outputs~\citep{wen2026toward}. Examining the supporting artifacts can therefore reveal data practices that the description alone does not communicate.

\toolname{} builds on description--behavior consistency analysis to assess practices specified jointly in code and natural-language instructions, distinguishing disclosure from contextual appropriateness: a mismatch may reveal handling that exceeds the stated purpose, but even an accurately described practice can violate contextual privacy norms, so we evaluate data practices against both the purpose communicated to the user and the norms governing the information's use.

\paragraph{Contextual Privacy in Language-Model Agents.}
Contextual Integrity (CI) characterizes privacy through information flows that conform to the norms of a social context~\citep{nissenbaum2004privacy,barth2006privacy}. This perspective is relevant to language-model assistants because information available to a model is not necessarily appropriate to disclose in its current task. Benchmarks covering conversational reasoning and persistent memory examine whether models make this distinction~\citep{mireshghallah2024secret,mireshghallah2026cimemories}.

For agents, contextual privacy must also be evaluated through their actions. PrivacyLens shows that correctly judging a disclosure scenario does not necessarily translate into appropriate disclosure during task execution~\citep{shao2024privacylens}. Similar work evaluates and mitigates disclosure in dynamic agent environments and examines information sharing within enterprise contexts~\citep{wang2025privacy,fu2026ciwork}. The unit of evaluation also matters. Reasoning traces and intermediate tool interactions can expose information that is absent from the final response, motivating assessments across the execution workflow~\citep{green2025leaky,hu2026toolprivacybench,ngong2026agentscope}. Moreover, privacy concerns extend beyond explicit disclosure: language models can derive sensitive personal attributes from otherwise available text~\citep{staab2024beyond}.

CI has become the dominant frame for privacy in language models, and these studies assess privacy through model judgments and generated behavior. We apply it to a static artifact rather than a running conversation, judging each extracted flow against both the purpose the skill communicates and the norms of the data's domain, to identify potentially inappropriate data handling before the skill is installed and given access to the user's personal information.

\begin{figure*}[t]
    \centering
    \includegraphics[width=\textwidth]{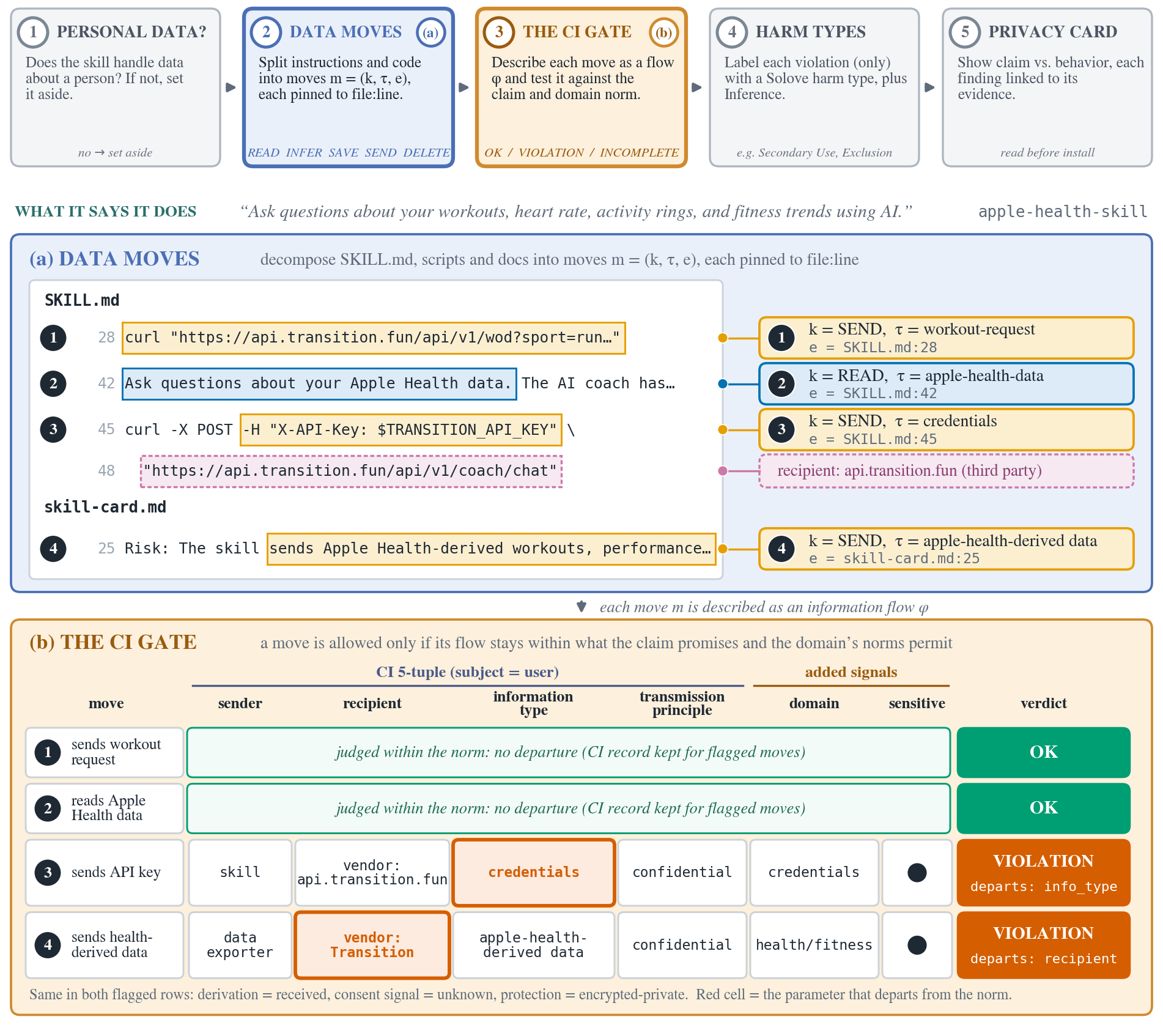}
        \caption{\toolname{} pipeline on a real skill (\texttt{apple-health-skill}), highlighting the pipeline stages with expanded data moves and the CI gate below.}
    \label{fig:skillprivus_full}
\end{figure*}
\section{SkillPrivus}
\label{sec:method}
We propose \toolname{} (Figure~\ref{fig:skillprivus_full}), a pipeline that analyzes a skill's instructions and supporting code to identify potential privacy harms before installation, without executing the skill.

\paragraph{Privacy Auditing.}
We focus on privacy harms that arise even when a skill performs its intended task. Our audit assesses whether the personal-data handling specified in its artifacts is appropriate for its stated purpose and context. This includes practices that exceed what the description communicates, as well as explicitly disclosed practices that violate contextual privacy expectations.

\paragraph{Audit Inputs and Scope.}
We start by defining a skill as a pair $s=(c,F)$, where $c$ is the natural-language \emph{claim} shown at install time and $F$ is the set of artifacts that encode its behavior. We distinguish the description that communicates the skill's intended functionality \emph{claim} from the instructions and code that specify how it operates $F$. The \texttt{SKILL.md} file may contain both the stated purpose and instructions that specify the skill's behavior.
The stated purpose provides a reference for assessing the data handling specified in the skill. From $F$ we recover a set of \emph{data moves} $\mathcal{M}(s)$, in which each $m=(k,\tau,e)$ is one operation $k$ over information of type $\tau$, grounded in source evidence $e$ (a file, a line range, and a backward slice).

A data move is the unit of analysis through which \toolname{} connects the behavior specified in a skill to a privacy assessment. Identifying an operation alone is insufficient for this assessment. For instance, a transmission cannot be judged without establishing what information is sent and where it is directed. The supporting evidence connects each operation to the information it handles. For code, we use backward program slicing~\citep{weiser1984slicing} to identify the statements on which the operation's relevant values depend. For natural-language instructions, we retain the passages specifying the operation. Each move records its source location and supporting evidence so that its interpretation can be inspected. 

\paragraph{Screening.}
Not all skills process personal data. \toolname{} therefore first examines each skill's artifacts for evidence that it may handle information about an individual. Skills without such evidence are excluded from subsequent privacy assessment. This step distinguishes skills with no identified personal-data handling from those whose data practices are assessed as appropriate.
We report the proportion of skills flagged for potential privacy harms among those that pass screening, together with the number retained from the full corpus.

\paragraph{Data moves.}
We use the operation vocabulary $k\in\{\textsc{Read},\textsc{Infer},\textsc{Save},\textsc{Send},
\textsc{Delete}\}$ to operationalize the broad notion of data
processing reflected in the GDPR~\citep{eu2016gdpr} and the NIST Privacy Framework~\citep{nist2020privacyframework}. These categories distinguish operations across the data lifecycle for the purposes of our audit. We explicitly distinguish \textsc{Infer} to capture the derivation of new personal attributes from available information, which may introduce privacy risks without additional collection or external disclosure.

\paragraph{Observed information flow.}
Each data move provides the basis for assessing whether the skill
handles information appropriately. \toolname{} constructs an \emph{observed information flow}  $\phi$ for each data move and evaluates it against the skill's claim and contextual privacy criteria. A departure from these criteria yields a \textsc{Violation} verdict and a harm type $h$ describing its potential privacy consequence. Otherwise, the move is marked \textsc{Ok}, or \textsc{Incomplete} when the evidence does not support a judgment. The resulting audit maps each skill to its assessed moves, $s \mapsto \{(m,v,h)\mid m\in\mathcal{M}(s)\}$, where $h=\varnothing$ for moves not classified as violations. Figure~\ref{fig:skillprivus_full} presents the pipeline as five stages --- screening, data moves, the CI gate, harm typing, and the \emph{Privacy Card} --- and Algorithm~\ref{alg:skillprivus} specifies the full procedure.

\begin{algorithm}[t]
\caption{\textsc{\toolname{}}: static privacy audit of an agent skill}
\label{alg:skillprivus}
\begin{algorithmic}[1]
\Require skill $s=(c,F)$; CI judge $J$; harm taxonomy $\mathcal{H}$
\Ensure findings $R$ (evidence-linked)
\State \textbf{if} $\neg\textsc{HandlesPersonalData}(s)$ \textbf{then return} $\varnothing$
       \Comment{screening: skills touching no personal data yield no findings}
\State $c \gets \textsc{Claim}(F);\quad B \gets \textsc{Behavior}(F)$
       \Comment{claim vs.\ behavior, from the same artifacts}
\State $\mathcal{M} \gets \textsc{ExtractMoves}(B)$
       \Comment{code and natural language, each pinned to a backward slice}
\State $R \gets \varnothing$
\For{each move $m=(k,\tau,e)\in\mathcal{M}$}
  \State $\phi \gets \textsc{ObservedFlow}(m)$ \Comment{CI 5-tuple + data-context signals}
  \State $v \gets J(\phi,\, c,\, \textsc{Domain}(m))$ \Comment{$v\in\{\textsc{Violation},\textsc{Ok},\textsc{Incomplete}\}$}
  \If{$v=\textsc{Violation}$}
    \State $h \gets \textsc{ClassifyHarm}(\phi,\,\mathcal{H})$
    \State $R \gets R \cup \{(m,\phi,v,h)\}$
  \EndIf
\EndFor
\State \Return $\textsc{PrivacyCard}(R)$
\end{algorithmic}
\end{algorithm}

% \paragraph{Contextual-integrity signals}
% \label{app:signals}

% For completeness, we list the data-context signals \toolname{} reads from the artifact to augment the classical Contextual Integrity 5-tuple \citep{nissenbaum2004privacy}. Each is derived statically from the skill's code and instructions, and each is grounded in source evidence like any data move.

% We additionally record \textbf{domain}, the social sphere governing the data (health, finance, correspondence, and so on), using an open vocabulary to accommodate heterogeneous skills; \textbf{sensitivity}, covering special-category data, credentials, and others' confided content, which should remain local absent explicit consent; \textbf{protection at the destination}, distinguishing private, local, shared, and public locations; \textbf{derivation}, our extension to CI for creating new sensitive attributes rather than transferring existing ones; and \textbf{consent signal}, artifact-level evidence that consent is sought, which informs interpretation but does not establish the user's actual decision.

\paragraph{Contextual-integrity signals.}
Table~\ref{tab:ci} lists the five data-context signals \toolname{} adds to the classical Contextual Integrity 5-tuple~\citep{barth2006privacy,nissenbaum2004privacy}. The 5-tuple captures \emph{how} information moves; the added signals supply the context needed to decide \emph{whether} a move is appropriate, with the consent signal used only as a modulator. Each is derived statically from the skill's code and instructions and grounded in source evidence like any data move.

\begin{table}[t]
\small\centering
\begin{tabular}{@{}l p{0.60\columnwidth}@{}}
\toprule
\textbf{Parameter} & \textbf{What it records} \\
\midrule
\multicolumn{2}{@{}l}{\emph{Contextual Integrity 5-tuple}}\\
Subject        & whose data the flow concerns \\
Sender         & the component that acts \\
Recipient      & where the data is directed \\
Information type & the attribute handled \\
Transmission   & condition on the flow (stay local, delete after use) \\
\addlinespace
\multicolumn{2}{@{}l}{\emph{Added data-context signals}}\\
Domain       & social sphere (health, finance, \ldots) \\
Sensitivity  & special-category data, credentials, confided content \\
Protection   & destination type: private, local, shared, public \\
Derivation   & creates a new sensitive attribute (our CI extension) \\
Consent signal & artifact-level evidence that consent is sought \\
\bottomrule
\end{tabular}
\caption{Parameters of the observed flow $\phi$: the classical CI 5-tuple plus five signals \toolname{} reads statically from the artifact. Each is grounded in source evidence.}
\label{tab:ci}
\end{table}

\paragraph{Deterministic checks.}
A small set of deterministic checks runs on top of the model's verdict for the cases a static reading can settle unambiguously, so that these do not depend on model judgment. A recipient at a loopback address is recorded as a local destination, a credential or secret is always marked sensitive, and personal data written to a world-readable temporary path and never deleted is recorded as a shared store rather than a transient local write. Each check encodes a norm already stated above and fires only on evidence visible at the cited line, and any move the model leaves unjudged is forced to \textsc{Incomplete} rather than passed silently, so coverage is complete and every verdict is either judged or explicitly marked unresolved.

% \paragraph{Judging appropriateness} 
% For each data move, \toolname{} constructs the observed flow $\phi$ from the five Contextual Integrity parameters~\citep{barth2006privacy,nissenbaum2004privacy}, the \emph{subject} the data concerns, the \emph{sender} that acts, the \emph{recipient} that receives it, the \emph{information type}, and the \emph{transmission principle}, the condition placed on the flow, such as keeping data local or deleting it after use. These parameters capture \emph{how} information moves, but deciding \emph{whether} a flow is appropriate needs context that classical CI leaves unstated, so \toolname{} adds five signals read statically from the artifact, the data \emph{domain}, its \emph{sensitivity}, the \emph{protection at the destination}, whether the move performs a \emph{derivation} that creates a new sensitive attribute (our extension to CI, which otherwise describes flows of attributes that already exist), and an observable \emph{consent signal} used only as a modulator.

\paragraph{The verdict.}
We call a flow appropriate when it stays within its \emph{norm}, the intersection of what the stated purpose entitles and what the flow's own domain conventionally entitles, in the sense of Contextual Integrity. We take the structure of CI, information flows described by parameters and judged against a context's norm, but not its method for \emph{deriving} that norm: rather than reading an entrenched social norm from established practice, we approximate it from the skill's own claim and the data's domain, an engineering adaptation we validate against human judgment (Section~\ref{sec:eval}) rather than a claim to have established the norm as CI prescribes. A move is flagged as a potential violation when it falls outside this norm on one of five root causes of violations: \textbf{recipient}, when data reaches an unwarranted recipient or shared store; \textbf{information type}, when processing exceeds the warranted data scope; \textbf{subject}, when another person's confided data is disclosed, retained, or profiled beyond contextual expectations; \textbf{access}, when permissions exceed the task's needs, even if unused in a particular execution; and \textbf{derivation}, when the skill infers a new sensitive attribute outside the claimed task. These root causes name specific ways a flow can be inappropriate, not a single risk score; Appendix~\ref{app:prompt} provides the rules.

Because a skill's author also defines its stated purpose, an overly broad claim could make any encoded behavior appear justified merely by describing it. To limit this, we impose a conservative policy: GDPR Article~9 special-category data~\citep{eu2016gdpr}, credentials, and another person's confided content must remain local unless the artifacts explicitly require user consent before disclosure. This locality requirement is an auditing choice, not a legal requirement derived from Article~9; an explicit consent requirement does not establish that consent was obtained at runtime. Other contextual norms are approximated by the model,
grounded in source evidence, and evaluated against human judgments (Section~\ref{sec:eval}). When the evidence cannot resolve a parameter required
for the verdict, the move is marked \textsc{Incomplete}, preserving uncertainty rather than treating it as appropriate. Appendix~\ref{app:example} provides a worked example, and Appendix~\ref{app:walkthrough} traces one skill through every stage of the pipeline.

\paragraph{Harm Taxonomy.}

We adapt Solove's taxonomy~\citep{solove2006taxonomy} to characterize the potential consequences of the violations identified by \toolname{}. The taxonomy distinguishes \emph{information collection} (surveillance, interrogation), \emph{information processing} (aggregation, identification, insecurity, secondary use, exclusion), \emph{information dissemination} (breach of confidentiality, disclosure, exposure, increased accessibility, blackmail, appropriation, distortion), and \emph{invasion} (intrusion, decisional interference). We additionally introduce \emph{Inference} as an explicit audit category for the inappropriate derivation of sensitive personal attributes. Deriving an attribute
does not itself constitute a violation; the judgment depends on the stated purpose and contextual privacy criteria. 

We exclude four categories whose assessment is not reliably supported by the evidence presented in the code alone. \emph{Insecurity} concerns whether stored data is adequately protected, which may depend on deployment conditions absent from the skill's files.
\emph{Blackmail} is a later act by whoever ends up holding the data, not the skill. \emph{Intrusion} concerns a disturbance of a person's solitude, while \emph{Decisional Interference} concerns interference with private choices. Both required real-world context the code never shows. 

The remaining thirteen categories are eligible harm labels: \emph{Surveillance}, \emph{Interrogation}, \emph{Aggregation}, \emph{Identification}, \emph{Secondary Use}, \emph{Exclusion}, \emph{Breach of Confidentiality}, \emph{Disclosure}, \emph{Exposure}, \emph{Increased Accessibility}, \emph{Appropriation}, \emph{Distortion}, and our \emph{Inference} extension. Each violation is characterized using the complete information flow, since a departure from the same contextual norm may have different consequences depending on the operation. Within information processing, \emph{Secondary Use} denotes uses beyond the skill's stated purpose, while \emph{Exclusion} denies individuals warranted opportunities to understand or participate in the handling of their data. Appendix Table~\ref{tab:harms} defines every harm label \toolname{} can assign and records the four Solove categories we exclude.

\paragraph{Harm typing and reporting.}
Each flagged move receives the most specific applicable category from the statically observable subset of Solove's taxonomy~\citep{solove2006taxonomy}, extended
with \emph{Inference} for newly derived sensitive attributes. The root cause of the violation constrains the applicable categories (Appendix~\ref{app:solove}); harm typing
characterizes the potential consequence without changing the verdict. 
% Findings are compiled into a \emph{Privacy Card} for non-technical readers~\citep{emaminaeini2020ask,kelley2009nutrition}, summarizing the data handled, destinations, and potential harms. Each finding links to its supporting data moves for inspection before installation (Appendix~\ref{app:card}). 

Findings are compiled into a \emph{Privacy Card} for non-technical readers~\citep{emaminaeini2020ask,kelley2009nutrition} (Appendix~\ref{app:card}), summarizing the data handled, destinations, and potential harms. The card contrasts the skill's stated purpose with its observed data practices and explains, in plain language, what could happen to the user's information. These risk descriptions convey plausible consequences of the flagged practices, rather than claims that harm has occurred or estimates of its likelihood. To support actionable decisions, the card also provides recommendations for reviewing the skill before installation, limiting exposure during use, and withdrawing access afterward. Depending on the identified practices, these may include checking external data transfers before connecting an account, reviewing sensitive data retained in temporary or shared locations, and revoking access when the skill is no longer needed. Recommendations guide user review and mitigation; they do not establish that a skill is safe or that all potential harms can be prevented. Each finding links to its supporting data moves and source evidence, allowing readers to inspect the basis for the reported risk before installation.

% \subsection{Severity and mitigation} %Not sure how to report this
% We assess potential harm severity using an impact-based
% rubric informed by NIST PRAM and adapted from the
% CNIL PIA methodology~\citep{,}.
% For each finding, the model describes a plausible
% consequence grounded in the source evidence and context,
% then assigns a severity level according to its impact
% and the difficulty of recovery. \emph{Low} denotes minor,
% readily recoverable consequences; \emph{Moderate},
% consequences recoverable with some difficulty;
% \emph{High}, serious consequences that are difficult
% to overcome; and \emph{Critical}, severe or enduring
% consequences that the individual may be unable to
% overcome. These ratings concern potential impact,
% not occurrence probability. Insufficient evidence
% yields an undetermined rating. The rubric and prompt
% appear in Appendix~\ref{app:severity-prompt}.

% Using the same underlying model, we generate mitigation
% guidance grounded in {}.
% Given the finding and its supporting implementation
% context, the model proposes a source-level change
% addressing the violated criterion while preserving
% the skill's stated purpose. Each recommendation
% identifies the affected location and a verification
% condition. Where external processing remains necessary,
% disclosure and user-control measures are recommended
% only if they address the identified concern.
% Unsupported repairs remain unresolved, with the missing
% information or functional dependency recorded.
% Appendix~\ref{app:mitigation-prompt} provides the
% strategy definitions and generation prompt.

\section{Experiments}

\subsection{Setup}
\paragraph{Skill corpus.}
We collected 726 \emph{skills from the wild} to study the data practices specified in publicly available agent skills. Our collection combines skills discovered through the ClawHub, SkillsMP, and skills.rest marketplaces, together with Anthropic's and NVIDIA's official collections and community-maintained repositories; Appendix Table~\ref{tab:corpus-sources} reports the distribution across sources. Screening set aside 118 skills with no personal-data handling, and 45 did not complete analysis because of malformed artifacts, leaving 563 skills with completed analyses over which we report prevalence.

\paragraph{Model.}
We implement \toolname{} with the open-weight \texttt{gpt-oss-120b} model.

\paragraph{Human annotation.}
To validate the judgments, we assembled a reference set for annotation. We selected skills with clear task descriptions that required little specialized domain knowledge and imposed a modest reading burden, prioritizing skills whose descriptions suggested personal-data handling while seeking coverage across all source collections. This process yielded 24 candidates, three of which were excluded by the pipeline's personal-data relevance gate (Appendix~\ref{app:relevance-gate}). The remaining 21 skills produced 339 data moves, reduced to 311 after removing 28 near-duplicates. Three doctoral students in STEM disciplines, recruited by email, each independently labeled every move as \textsc{Violation}, \textsc{Appropriate}, or \textsc{Unresolvable}, the last reserved for cases where the evidence does not support a judgment, spending approximately two hours; a 20-move pilot checked that the guidelines were usable. For each move, annotators inspect the skill's claim, the supporting source evidence, and the information flow reconstructed by \toolname{}, but not the system's inferred norm or its final assessment (Figure~\ref{fig:humanUI2}).

\subsection{Evaluation}
\label{sec:eval}
We validate \toolname{}'s judgments against the three annotators before reporting prevalence. Because appropriateness is a subjective judgment, we do not treat any single label as ground truth. We instead ask whether \toolname{} agrees with a human about as often as two humans agree with each other. For every pair of raters, two annotators or an annotator and \toolname{}, we report percentage agreement and Cohen's $\kappa$, excluding moves either rater marked \textsc{Unresolvable}, and we average over the three pairs; across all three annotators we also report Krippendorff's $\alpha$. The three annotators agreed 83.4\% of the time (mean $\kappa=0.478$, $\alpha=0.380$), and \toolname{} agreed with them 83.2\% of the time ($\kappa=0.346$), almost as often as they agreed with one another.  These $\kappa$ values are moderate for two reasons. First, almost all moves are appropriate, and $\kappa$ is known to fall when one label dominates, even when raters largely agree~\citep{feinstein1990high}. Second, annotators disagree on genuinely borderline cases, which reflects the subjectivity at the heart of the task, not labeling error.

\begin{figure*}[!t]
\centering
\definecolor{barblue}{HTML}{4C72B0}
\pgfplotsset{
violbar/.style={
xbar, clip=false,
ytick=data, y dir=reverse, enlarge y limits=false,
axis x line*=bottom, axis y line*=left, y axis line style={draw=none},
ytick style={draw=none},
xmajorgrids, grid style={gray!30, line width=0.3pt},
y tick label style={align=right},
tick label style={font=\scriptsize},
nodes near coords, point meta=explicit symbolic,
every node near coord/.append style={anchor=west, xshift=1.5pt, text=black, font=\scriptsize},
every axis plot/.append style={fill=barblue!85, draw=barblue!85!black, line width=0.3pt},
},
}

\begin{minipage}[c]{0.56\textwidth}
\centering
\begin{tikzpicture}
\begin{axis}[violbar, width=0.86\linewidth, y=0.60cm, bar width=9pt,
xmin=0, xmax=68, xtick={0,20,40,60}, xticklabels={0,20,40,60\%},
ymin=0.3, ymax=8.7,
yticklabels={Exclusion,Secondary Use,Breach of Conf.,Incr. Accessibility,Disclosure,Inference,Surveillance,Aggregation}]
\addplot coordinates {
(53.9,1) [1{,}156]
(25.7,2) [552]
(6.6,3) [142]
(4.2,4) [91]
(3.8,5) [82]
(2.9,6) [63]
(2.5,7) [54]
(0.2,8) [4]};
\end{axis}
\end{tikzpicture}
\par\vspace{3pt}
{\footnotesize (a) Harm type}
\end{minipage}
\hfill
\begin{minipage}[c]{0.40\textwidth}
\centering
\begin{tikzpicture}
\begin{axis}[violbar, width=0.90\linewidth, y=0.495cm, bar width=6pt,
xmin=0, xmax=78, xtick={0,20,40,60}, xticklabels={0,20,40,60\%},
ymin=0.3, ymax=5.7,
yticklabels={Information type,Recipient,Subject,Derivation,Access}]
\addplot coordinates {
(54.2,1) [54.2\%]
(33.7,2) [33.7\%]
(6.5,3) [6.5\%]
(3.2,4) [3.2\%]
(2.4,5) [2.4\%]};
\end{axis}
\end{tikzpicture}
\par\vspace{2pt}
{\footnotesize (b) Root cause of violations}
\par\vspace{8pt}
\begin{tikzpicture}
\begin{axis}[violbar, width=0.90\linewidth, y=0.495cm, bar width=6pt,
xmin=0, xmax=48, xtick={0,10,20,30,40}, xticklabels={0,10,20,30,40\%},
ymin=0.3, ymax=5.7,
yticklabels={Read,Send,Save,Infer,Delete}]
\addplot coordinates {
(34.3,1) [735]
(31.7,2) [679]
(23.5,3) [503]
(9.6,4) [206]
(1.0,5) [21]};
\end{axis}
\end{tikzpicture}
\par\vspace{2pt}
{\footnotesize (c) Operation carrying it}
\end{minipage}

\caption{Where the 2{,}144 flagged data moves come from. Each panel partitions the same 2{,}144 moves and is drawn as a percentage of flagged moves. (a) The harm type assigned to the move and (c) the operation that carried it are labeled with move counts; (b), the root cause of the violation, is labeled with percentages.}
\label{fig:results}
\end{figure*}
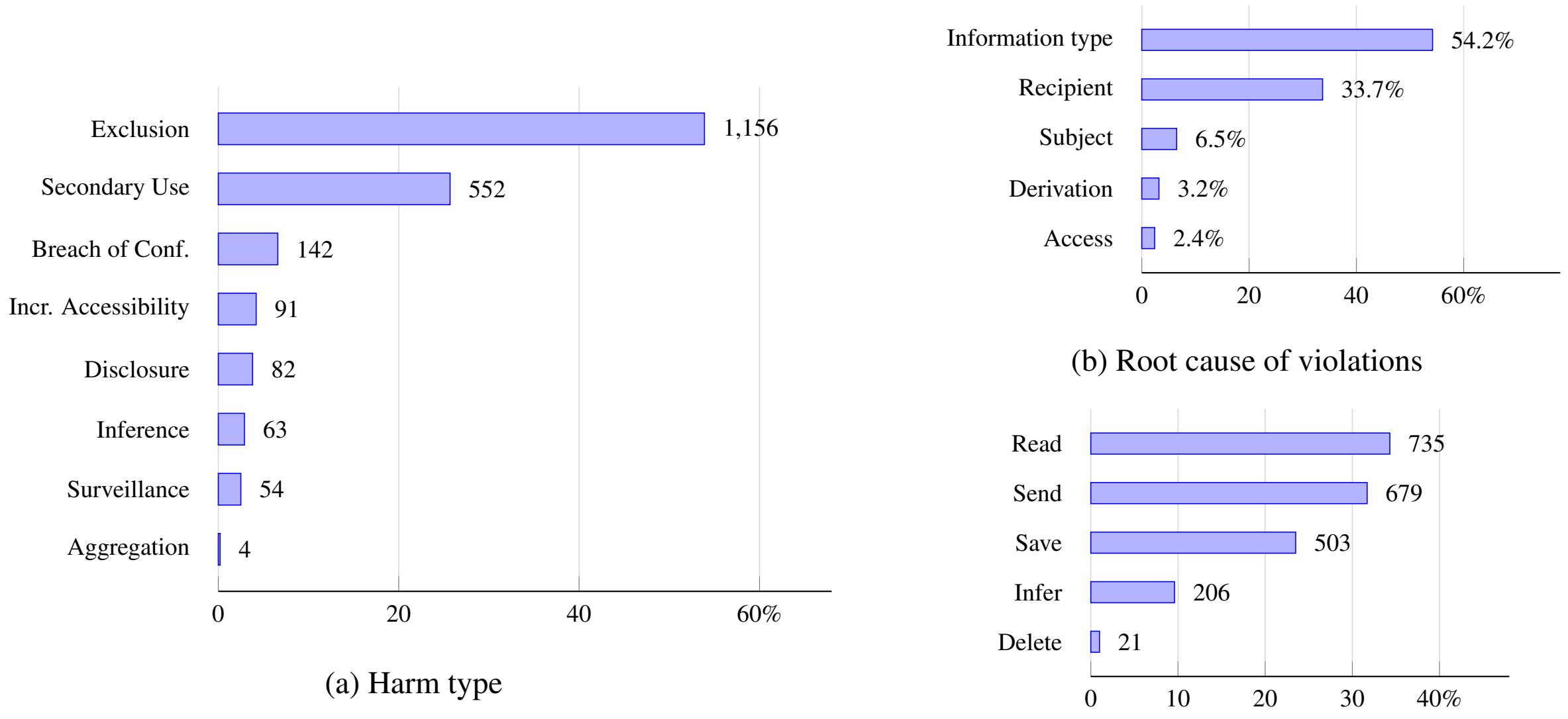

\section{Results}
\label{sec:results}
Over the 563 skills with completed analyses, \toolname{} extracted 22{,}158 data moves and flagged 2{,}144 (9.7\%) as potential privacy violations. More than half, 307 of 563 skills (54.5\%), contained at least one data move flagged as inconsistent with the skill's stated purpose or contextual privacy norms.

\paragraph{Most violations come from the kind of data and where it goes.}
Figure~\ref{fig:results}(b) breaks each flagged move down by its root cause, the part of the flow that made it inappropriate. The two leading causes are the \emph{information type}, where a skill handles data beyond what its stated purpose covers (54.2\%), and the \emph{recipient}, where data goes to a destination the context does not support (33.7\%). The other causes are far less common: the \emph{subject} of the data (6.5\%), an unwarranted \emph{derivation} of a new attribute (3.2\%), and excess \emph{access} (2.4\%). Separately, Figure~\ref{fig:results}(c) shows that flags fall mostly on routine read, send, and save operations rather than inference, so these violations arise within ordinary data handling.

\paragraph{The most common harm is handling more data than the description admits.}
Among the 2{,}144 flagged data-movement records,
\emph{Exclusion} is the most frequent harm label, accounting
for 1{,}156 records (53.9\%). In our operationalization,
this label denotes accessing or retaining data beyond what
the skill's description discloses. \emph{Secondary Use}
follows with 552 records (25.7\%), then \emph{Breach of
Confidentiality} with 142 (6.6\%), \emph{Increased
Accessibility} with 91 (4.2\%), and \emph{Disclosure}
with 82 (3.8\%). Figure~\ref{fig:results}(a) shows the
distribution across all observed harm categories.
Each flagged record contributes one harm label.

%Per skill
% At the skill level, \emph{Exclusion} is the most frequent harm, in 199 of 563 skills (35.3\%), where a skill accesses or retains more data than its description discloses. \emph{Secondary Use} follows in 138 skills (24.5\%), where data is used or transferred beyond the original purpose, then \emph{Breach of Confidentiality} in 61 skills (10.8\%), \emph{Increased Accessibility} in 42 (7.5\%), and \emph{Disclosure} in 40 (7.1\%). Figure~\ref{fig:results}(a) shows the corresponding distribution across the 2{,}144 flagged moves. The dominant patterns concern the scope, use, and destination of data, showing how potential privacy harms can arise within ordinary skill functionality without requiring an attack.

\paragraph{Security screening does not catch these.}
NVIDIA's Verified Skills are scanned and signed by its pipeline, which runs the \textsc{SkillSpector} security scanner for vulnerabilities and malicious intent, before publication~\citep{nvidia_skillspector}. On the 37 such skills in our corpus, \toolname{} flagged potential privacy-of-use violations in 24 (Appendix~\ref{app:nvidia}).

\section{Discussion}
The dominance of information-type and recipient departures among flagged moves (Figure~\ref{fig:results}(b)) points to the scope and destination of data, not sensitivity alone.

\textbf{The install-time description cannot be trusted.} Across our corpus it understated what the skill does for most privacy-relevant skills (54.5\%), so any check that reads only the claim, as app-store review and permission prompts largely do~\citep{gorla2014chabada,qu2014autocog}, misses privacy of use. The behavior has to be read from the skill itself, before it runs.

\textbf{Sensitivity is the wrong filter.} What goes wrong is who receives the data and what kind of data the skill reaches, not one especially sensitive field. A filter that ranks by sensitivity alone lets such moves through: sending data to the wrong recipient is a violation whatever the data's sensitivity, the distinction Contextual Integrity is built to draw.

\textbf{Security review is not enough on its own.} Every harm we find is honest-but-negligent, a skill using access it was granted for more than the task needed, not an attack. There is no adversary to catch, so reading a skill's behavior before installation adds a line of defense that security tooling, aimed at a different threat, is not built to provide.

\textbf{Prevention belongs with the skill's creator.} Nothing stops a creator from running \toolname{} as a pre-release scan and fixing privacy-of-use violations before a skill ships, and a \emph{Privacy Card} published with each skill turns the same evidence into a disclosure a user can read at install time. Appropriateness then becomes something a skill is authored with, not audited for after the fact; these applications remain to be evaluated.

\section{Conclusion}
We introduced \toolname{}, a static framework that assesses the privacy appropriateness of an agent skill before installation by reading its behavior against its stated purpose. The underlying problem, privacy of use, has lacked an agreed name, documented cases, and a measure; this paper supplies each. Across 563 skills, \toolname{} flagged potential violations in 54.5\%, most often handling data beyond the stated scope or putting it to a secondary use, and it surfaced contextual concerns in skills that had already passed security scanning. These findings show that privacy review must examine a skill's ordinary functionality, not only its security, and that an install-time description cannot be trusted to disclose what the skill does. Its judgments track human annotators to near the inter-annotator ceiling; broader expert validation and runtime evaluation remain to establish when flagged behaviors lead to realized harm.

%%%
% We introduced \toolname{}, a static framework for assessing privacy appropriateness in agent skills before installation. Across 563 skills with completed analyses, it flagged potential violations in 54.5\%, most commonly involving data handling beyond the stated scope and secondary use. These findings suggest that privacy review must examine ordinary functionality alongside security threats. Agreement with human annotators provides initial support for the approach, while broader validation and runtime evaluation remain necessary to establish when flagged behaviors lead to realized harms.

\section{Limitations and Future Work}
Reading a skill statically recovers what it \emph{would} do, so it cannot confirm that a flow occurs on a given run, and behavior decided at runtime is harder to pin down; we therefore report potential violations, each backed by evidence a reader can check, rather than final verdicts. A dedicated audit of extraction fidelity is left to future work.

Our validation covers a reference set of 311 moves from 21 skills labeled by three annotators, and contextual appropriateness is contested, as $\kappa=0.346$ reflects; larger expert-annotated sets would sharpen the evaluation across models, marketplaces, and domains. The domain-norm half of the verdict is at present the model's own judgment, which we check against human labels rather than assume correct; grounding it in a written set of domain norms, or a model trained on cases experts have already ruled on, would put that half on a footing as clear as the legal floor.

Screening selects skills from the description and a slice of the code, so a skill that hides all its personal-data access in code the description never mentions could be set aside by mistake; that kind of deliberate hiding is the undisclosed-access problem security tools already target, outside our scope. Finally, our corpus is a snapshot of public marketplaces at one moment, and we report prevalence over the privacy-relevant skills rather than the whole marketplace.

The \emph{Privacy Card}'s recommendations are guidance, not validated remedies: we do not evaluate whether following them reduces the identified harms, and their effect may depend on the skill's implementation, the user's control over its permissions, and the practices of external services. Future work could test their feasibility and effect through user studies.

\section{Ethical Considerations}
The motivation behind our work is to protect users’ privacy when they install skills and help prevent the use of their personal data without consent. Our study examines publicly available skill instructions and code without executing the skills or accessing users’ private accounts or records. The analysis identifies potential privacy harms supported by these artifacts; it does not establish that a flagged behavior occurred during use or caused harm to an individual.

Automated privacy assessments may mischaracterize legitimate functionality or overlook relevant context. We therefore report findings as potential violations, retain supporting source evidence, and distinguish unresolved cases from assessed practices. Findings should not be interpreted as evidence of malicious intent, legal noncompliance, or a definitive ranking of developers or providers. Similarly, the absence of findings does not certify a skill as safe. Our analysis of security-screened skills addresses contextual privacy questions and should not be understood as demonstrating that existing verification processes failed their stated objectives.

Contextual privacy expectations may vary across users, communities, and domains. The norms approximated by our framework and the judgments of three annotators do not represent all affected populations. The Privacy Card is intended to support informed review rather than replace users’ judgment or shift responsibility for privacy protection entirely onto them. Its recommendations are guidance, and their effectiveness in preventing or reducing harm has not been established.


\bibliography{custom}
\onecolumn

\appendix

\section{Corpus Sources}

\begin{table}[ht]
\centering
\begin{tabular}{lr}
\hline
\textbf{Source} & \textbf{Skills} \\
\hline
ClawHub     & 227 \\
SkillsMP    & 231 \\
skills.rest & 99  \\
Anthropic   & 19  \\
NVIDIA      & 50  \\
Glebis      & 97  \\
ClaudeClaw  & 3   \\
\hline
\textbf{Total} & \textbf{726} \\
\hline
\end{tabular}
\caption{Source distribution of the 726 skills from the wild
in our corpus. Sources denote collection origins rather than
distinct authorship.}
\label{tab:corpus-sources}
\end{table}

\section{Solove's Taxonomy}
\label{app:solove}

Table~\ref{tab:harms} gives the complete label set summarized in
Section~\ref{sec:method}. 

\begin{table}[t]
\small\centering
\begin{tabular}{@{}l p{0.56\columnwidth}@{}}
\toprule
\textbf{Harm label} & \textbf{Meaning in our audit} \\
\midrule
\multicolumn{2}{@{}l}{\emph{Information collection}}\\
Surveillance   & watching or monitoring a person's activity \\
Interrogation  & pressuring a person to divulge information \\
\addlinespace
\multicolumn{2}{@{}l}{\emph{Information processing}}\\
Aggregation    & combining data to learn more than any piece reveals \\
Identification & linking data to a specific person \\
Secondary Use  & use beyond the skill's stated purpose \\
Exclusion      & denies a person knowledge of, or a say in, how their data is handled \\
\addlinespace
\multicolumn{2}{@{}l}{\emph{Information dissemination}}\\
Breach of Confidentiality & shares something told in confidence (e.g., a contact's message, a cancellation reason) with another party \\
Disclosure              & reveals true information about a person to others \\
Exposure                & reveals a person's private physical or emotional details \\
Increased Accessibility & makes data reachable by more parties than the context warrants (e.g., writing it to a world-readable path) \\
Appropriation           & uses a person's identity for another's ends \\
Distortion              & spreads false or misleading information about a person \\
\addlinespace
\multicolumn{2}{@{}l}{\emph{Our extension}}\\
Inference & derives a new sensitive attribute the task did not warrant (e.g., health or behavior patterns from activity data) \\
\midrule
\multicolumn{2}{@{}l}{\emph{Excluded: a static reading of code cannot settle these}}\\
Insecurity              & adequacy of protection depends on deployment, not the files \\
Blackmail               & a later act by whoever holds the data, not the skill \\
Intrusion               & needs real-world context on a person's solitude \\
Decisional Interference & needs real-world context on private choices \\
\bottomrule
\end{tabular}
\caption{Harm labels, adapted from Solove's taxonomy~\citep{solove2006taxonomy}. We type each violation with the most specific applicable label. \emph{Inference} is our addition; the four excluded categories require runtime or real-world context the code does not provide.}
\label{tab:harms}
\end{table}

\section{Worked examples}
\label{app:example}

The three cases below are real skills from our corpus, judged by the pipeline of Section~\ref{sec:method}.
They show three different ways a skill goes wrong: sending data to an outside recipient, reaching more kinds
of data than the claim covers, and leaving data where anyone can read it. Every finding cites the exact
line, so a reviewer can check it against the source.

\begin{tcolorbox}[colback=gray!5,colframe=gray!55!black,boxrule=0.4pt,left=6pt,right=6pt,top=4pt,bottom=4pt]
\small
\textbf{1. \texttt{apple-health} --- health data leaves for a third party.} The skill claims to let you
``talk to your Apple Health data.'' Reading that data on the user's own machine is exactly its job, and
\textsc{\toolname{}} clears every such move. But the skill also \texttt{POST}s the user's health question, and
its API key, to \texttt{api.transition.fun}, a vendor the description never names (\texttt{SKILL.md}:45--48),
and does the same with a workouts request (\texttt{SKILL.md}:63--64). The flow departs on the
\emph{recipient}: health data and a credential go to an outside service the purpose does not justify.
Verdict: four violations, all \emph{Secondary Use}.
\end{tcolorbox}

\begin{tcolorbox}[colback=gray!5,colframe=gray!55!black,boxrule=0.4pt,left=6pt,right=6pt,top=4pt,bottom=4pt]
\small
\textbf{2. \texttt{zoom} --- one skill, four kinds of harm.} The skill claims to create and manage Zoom
meetings and access recordings. Along the way it saves meeting metadata, including the participants, to a
local file (\emph{Breach of Confidentiality}, subject departure, \texttt{SKILL.md}:12); summarizes the
recordings when the claim only said it would \emph{access} them (\emph{Inference}, \texttt{SKILL.md}:13);
reads the user's timezone, which the claim never mentions (\emph{Surveillance}, \texttt{SKILL.md}:15); and
ships OAuth credentials and API calls to Zoom's servers from its scripts (\emph{Secondary Use},
\texttt{scripts/zoom\_meetings.py}:100--191). Twelve violations across four harm types in a single, ordinary
skill.
\end{tcolorbox}

\begin{tcolorbox}[colback=gray!5,colframe=gray!55!black,boxrule=0.4pt,left=6pt,right=6pt,top=4pt,bottom=4pt]
\small
\textbf{3. \texttt{chrome-history} --- browsing data left in the open.} The skill claims to let you query
your Chrome history in natural language. To do so it copies the entire Chrome history database to a
world-readable \texttt{/tmp} path and never deletes it (\emph{Increased Accessibility}, recipient departure
to a shared store, \texttt{chrome\_history\_query.py}:300), and it infers the user's device type from the
records, a fact the claim never asked for (\emph{Inference}, \texttt{chrome\_history\_query.py}:152). Nothing
is sent anywhere, yet the browsing history is left where any other process on the machine can read it.
\end{tcolorbox}

\section{Pipeline walkthrough: \texttt{apple-health}}
\label{app:walkthrough}
\begin{figure*}[ht]
    \centering
    \includegraphics[width=\textwidth]{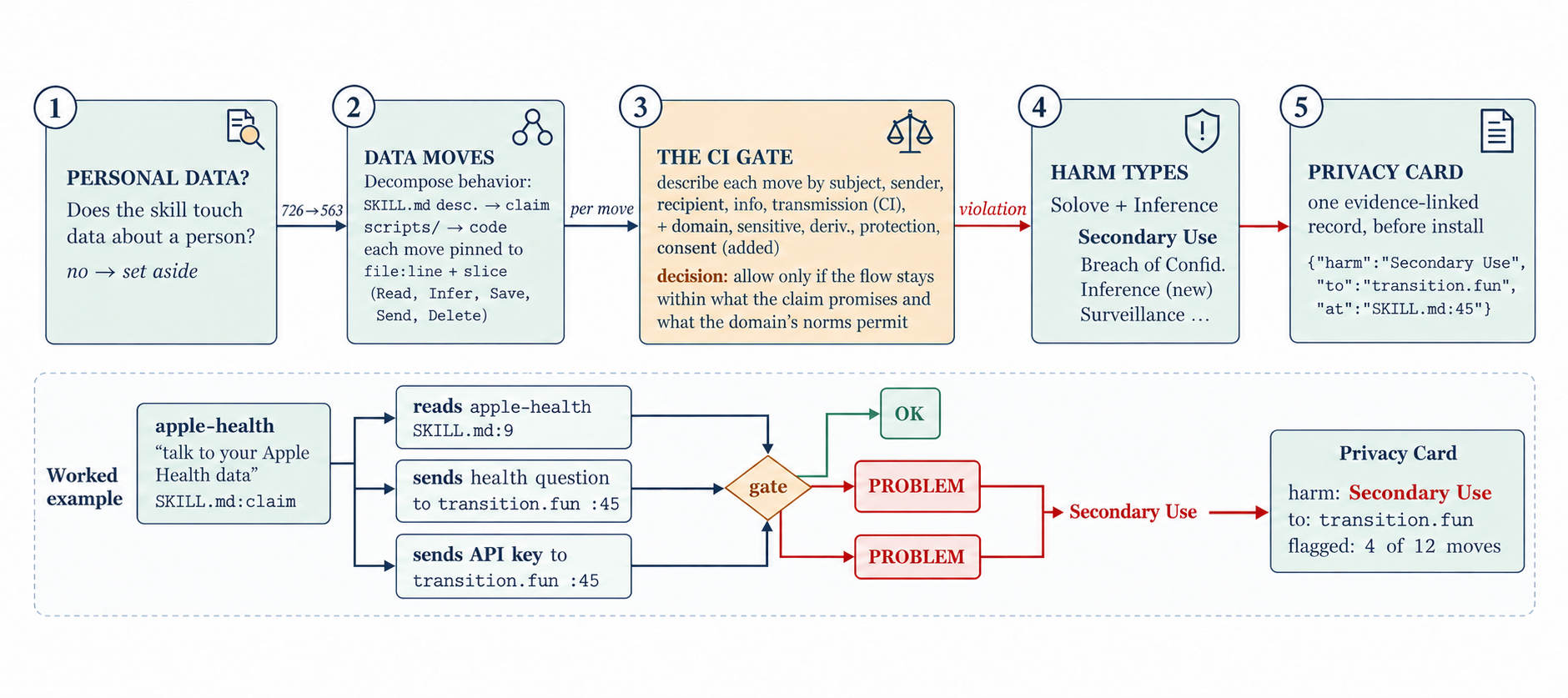}
    \caption{Worked example of \toolname{}. This example illustrates
    a secondary-use finding.}
    \label{fig:skillprivus}
\end{figure*}

Figure~\ref{fig:skillprivus} summarizes the audit of this skill. Figures~\ref{fig:move-to-ci}--\ref{fig:harm-typing} follow the first
worked example of Appendix~\ref{app:example} through every stage of
\toolname{}. The skill is an ordinary marketplace skill whose claim is
``talk to your Apple Health data''; no part of the audit requires running
it. We show the appropriateness judgment at the granularity of one move and of
the whole skill, and then harm typing. Each figure highlights only a subset of the skill's flagged moves so that the panels stay legible, and the counts therefore differ between figures; the largest count shown, the 11 moves in Figure~\ref{fig:ci-gate}, is the complete set of violations \toolname{} reports for this skill.

Extraction is deliberately permissive: it records what the artifacts
instruct, not whether the instruction is acceptable. The ledger therefore
contains appropriate and inappropriate moves side by side, and every entry
carries the source line a reviewer can check.

\begin{figure*}[ht]
  \centering

  \includegraphics[width=\textwidth]{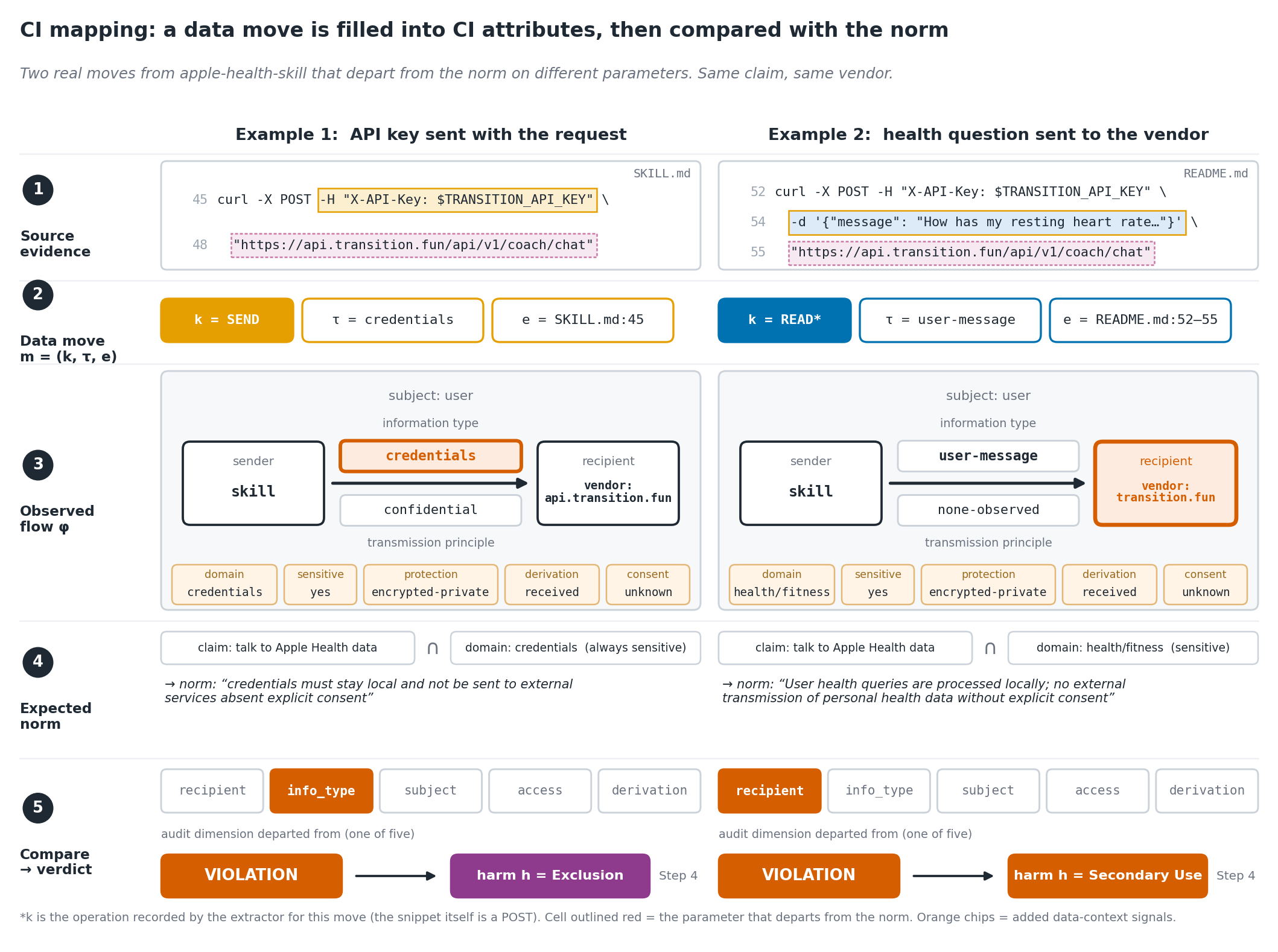}
  \caption{\textbf{For a single move.} Two moves of the same skill
  that fail on different CI parameters. Row~1 gives the source evidence;
  row~2 the extracted move $m=(k,\tau,e)$; row~3 the observed flow $\phi$,
  the move filled into the five CI parameters together with the added
  data-context signals; row~4 the expected norm, obtained by intersecting
  the skill's claim with the norms of the data domain; row~5 the comparison,
  which names the single root cause and hands the
  violation to Step 4. Example~1 departs on \texttt{info\_type}, a
  credential leaving the machine, and is typed \emph{Exclusion}; Example~2
  departs on \texttt{recipient}, a health question reaching an undisclosed
  vendor, and is typed \emph{Secondary Use}.}
  \label{fig:move-to-ci}
\end{figure*}

The norm is never written by hand. It is the conjunction of what the claim
entitles the skill to do with what the data domain permits, which is why
the same operation can be appropriate in one skill and a violation in
another.

\begin{figure*}[ht]
  \centering
  \includegraphics[width=\textwidth]{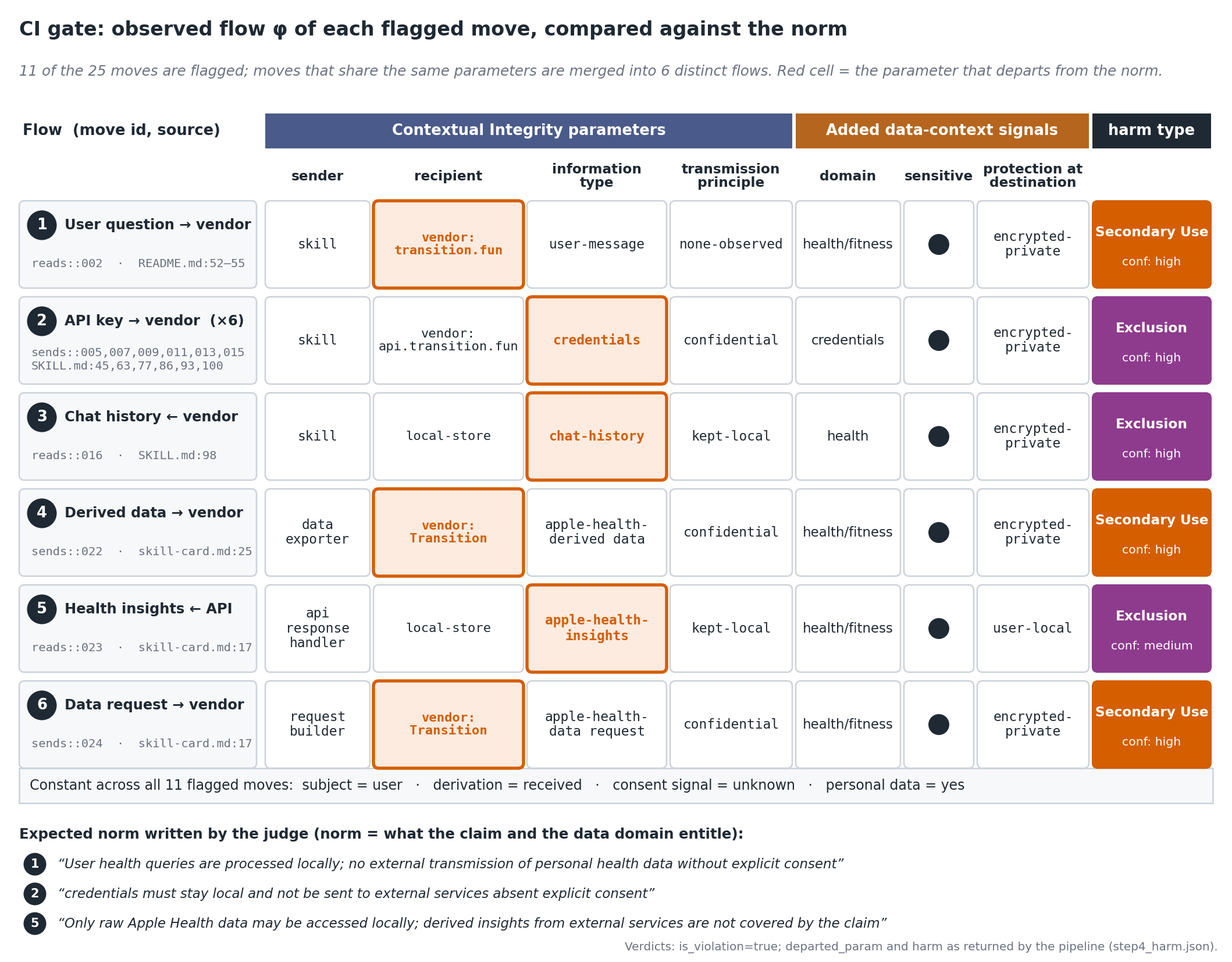}
  \caption{\textbf{Across the whole skill.} The 11 flagged moves
  reduce to six distinct observed flows once moves with identical CI
  parameters are merged; the API-key flow alone accounts for six of them.
  Columns give the five CI parameters, then the added data-context signals,
  then the harm type returned by Step 4. The outlined cell in each row is
  the parameter that departs from the expected norm: four flows depart on
  information type and two on recipient. The attributes beneath the table
  are constant across all 11 flagged moves, and the norms quoted at the
  foot are the judge's own statements of what the claim and the domain
  entitle.}
  \label{fig:ci-gate}
\end{figure*}

\begin{figure*}[ht]
  \centering
  \includegraphics[width=\textwidth]{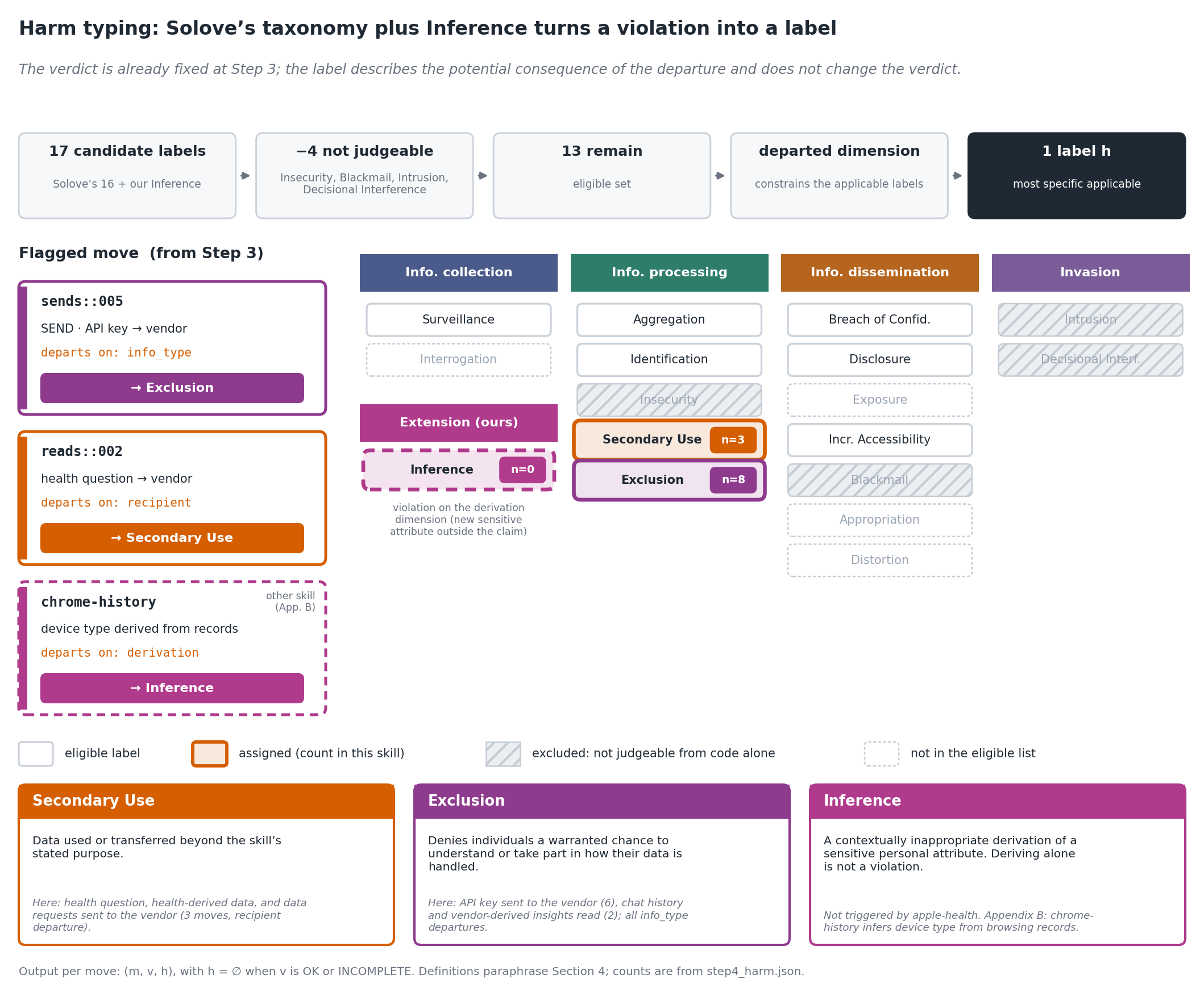}
  \caption{\textbf{Harm typing.} Typing labels a violation that
  Step 3 has already established; it never changes the verdict. The
  eligible set is Solove's sixteen categories plus our \emph{Inference}
  extension, less the four a static reading of code cannot settle
  (Insecurity, Blackmail, Intrusion, Decisional Interference), leaving
  thirteen. The root cause of the violation constrains which of the thirteen
  apply, and the most specific applicable label is assigned. Here this
  gives \emph{Exclusion} for the eight information-type departures and
  \emph{Secondary Use} for the three recipient departures.
  \emph{Inference} is not triggered by this skill; the third panel shows
  the \texttt{chrome-history} skill of Appendix~\ref{app:example}, which
  triggers it by deriving a device type from browsing records.}
  \label{fig:harm-typing}
\end{figure*}

\section{Privacy Card}
\label{app:card}

The findings for a skill are compiled into a \emph{Privacy Card}: a compact, evidence-linked summary a
non-technical user can read before installing. It names what data the skill handles, where that data goes,
and which harms were detected, and every line traces back to the data moves that produced it. Below is the
card for the \texttt{apple-health} skill of Appendix~\ref{app:example}, the same skill worked through in
Figure~\ref{fig:skillprivus_full}. Like the walkthrough figures, the card lists only a subset of the skill's moves; the complete audit flags 11 of the 25 moves extracted for this skill (Appendix~\ref{app:walkthrough}). The banner also counts user-facing practices rather than individual moves: the four flagged moves listed below collapse into the two practices the card names.

\begin{tcolorbox}[colback=gray!5,colframe=gray!55!black,boxrule=0.4pt,left=6pt,right=6pt,top=4pt,bottom=4pt]
\small
\textbf{Privacy Card: \texttt{apple-health}}\\[2pt]
\emph{Claim.} Talk to your Apple Health data --- ask about your workouts, heart rate, and fitness trends.\\[3pt]
\textbf{Data handled.} Apple Health records (workouts, heart rate, activity); the user's health questions;
the skill's API key.\\[3pt]
\textbf{Where it goes.} Reading your health data stays on your machine, but your health questions and
date-range requests, together with the API key, are sent to \texttt{api.transition.fun}, an outside vendor
the description never names.\\[3pt]
\textbf{Harms detected.}
\begin{itemize}\setlength{\itemsep}{1pt}
\item \emph{Secondary Use}: sends a health question to the vendor (\texttt{SKILL.md}:45--48).
\item \emph{Secondary Use}: sends the API key alongside it (\texttt{SKILL.md}:45--48).
\item \emph{Secondary Use}: sends a workouts request, and the key again, to the vendor
  (\texttt{SKILL.md}:63--64).
\end{itemize}
\textbf{Verdict.} 4 potential violations across 12 data moves. Inspect any finding at its cited line.
\end{tcolorbox}

% \begin{figure}[t]
%     \centering
%     \includegraphics[width=0.9\columnwidth]{figures/privacyCard.png}
%     \caption{\toolname{} Privacy Card.}
%     \label{fig:humanUI1}
% \end{figure}

\section{The appropriateness prompt}
\label{app:prompt}

This appendix shows the prompt that decides each verdict,
the step labeled $J$ in Algorithm~\ref{alg:skillprivus}.
The judge receives the skill's claim and a batch of
extracted data moves, and for each move it fills the
Contextual Integrity flow, names the data's domain in
its own words, writes down what the claim and the domain
together entitle, and then compares the observed flow
against that norm. It returns one JSON record per move.
Prompt templates use \texttt{\{VARIABLE\}} placeholders
that the pipeline substitutes per call:
\texttt{\{skill\}} and \texttt{\{claim\}} come from the
skill, and \texttt{\{moves\_block\}} is the list of moves
to judge. The prompt below is condensed for readability;
the full text, with the complete rule set and worked
examples, is in the released codebase under
\texttt{prompts/}.

The first box below is the prompt template; the second
is a real call and its response on the
\texttt{chrome-history} skill of
Appendix~\ref{app:example}.

\begin{tcblisting}{
    listing only,
    enhanced,
    breakable,
    width=\linewidth,
    colback=gray!5,
    colframe=black,
    coltitle=white,
    colbacktitle=gray!70,
    fonttitle=\bfseries,
    title={Appropriateness Judge Prompt (condensed)},
    title after break={Appropriateness Judge Prompt (continued)},
    boxrule=0.8pt,
    sharp corners,
    boxsep=0.5mm,
    left=0.5mm,
    right=0.5mm,
    top=0.5mm,
    bottom=0.5mm,
    before skip=6pt,
    after skip=6pt,
    listing options={
        basicstyle=\scriptsize\ttfamily,
        breaklines=true,
        breakatwhitespace=false,
        columns=fullflexible,
        keepspaces=true,
        showstringspaces=false
    }
}
You are a privacy auditor judging data moves already found in an AI "skill".
This is privacy-of-USE: there is NO attacker. A move can be perfectly in-scope
and STILL be a violation.

You judge with Contextual Integrity (CI): a flow is described by 5 parameters
and VIOLATES when its actual parameters DEPART from the norm. The norm = what
the skill's CLAIM entitles AND what the data's DOMAIN conventionally entitles.
For each move, in order:

  (0) GATE: does this flow involve a PERSON's data? Stop only when the data is
      neither personal nor sensitive-class. Credentials and another person's
      confided content are ALWAYS judged in full, even if not about a person.
  (1) fill the 5 CI parameters of the actual flow (subject, sender, recipient,
      information type, transmission principle);
  (2) name the DOMAIN in your own words (open-ended: health, finance, messaging,
      location, credentials, ...) and set `sensitive` = must this data stay
      local absent explicit consent? (true for a GDPR-Article-9 special
      category, credentials/secrets, or another person's confided content);
  (3) write the EXPECTED NORM = what the claim AND the domain entitle;
  (4) decide `is_violation` by comparing the actual flow to that norm, plus a
      derivation check.

The skill's CLAIM (one input to the norm; the domain is the other):
  skill:  {skill}
  claim:  {claim}

A move is a VIOLATION if the flow departs from the norm on any parameter, OR it
derives a new sensitive attribute. Set `departed_param` to the one that departed:
  recipient  the actual recipient is one the norm does not entitle (an external
             vendor, a public place, another person, or a shared store). A
             world-readable path (/tmp) holding personal data with no cleanup is
             a shared store. But a query the user composed FOR the service the
             skill exists to call is the claimed purpose, not a leak.
  info_type  the data reaches or keeps MORE than the claim's words cover (claim
             "transcribe" but the code reads the raw audio blob).
  subject    the actual subject is another person, and their confided data is
             sent out, copied into a new persistent artifact, or profiled.
  access     the skill acquires access broader than its purpose needs (OAuth
             scopes or file access reaching data the task never requires).
  derivation the move derives a sensitive attribute OUT of the claim's domain (a
             "transcribe" skill that outputs a mental-health label). Producing
             the claim's own stated analysis is in-domain and NOT a violation.

Two things that do NOT excuse a violation: PURPOSE ("sending is what this skill
does" never clears a recipient/info_type/subject departure) and CONSENT (an
observable consent signal only modulates; it never turns a real departure into
"false"). SENSITIVE DATA OVERRIDES A PERMISSIVE CLAIM: special-category data,
credentials, or another person's confided content stays local absent explicit
consent, even if the claim says it may leave.

Flows that MATCH the norm are NOT violations: a local endpoint (localhost, a
local port, the user's own files) as recipient; detecting sensitive spans in
order to redact them; outputting data this skill already anonymized. Mark
`incomplete` only when a CI parameter genuinely cannot be determined.

## The data moves to judge
{moves_block}

## Output: a JSON array, one element per move
[
  {
    "move_id": "<the exact move_id>",
    "reasoning": "<deciding line; domain + sensitivity; which parameter departs>",
    "ci": {"subject":"...","sender":"...","recipient":"...",
           "how_protected":"encrypted-private | user-local | shared | public",
           "info_type":"...",
           "transmission_principle":"kept-local | confidential | anonymized |
                                     not-stored | aggregated | none-observed",
           "domain":"<your own words>","sensitive": true | false,
           "personal_data":"yes | no | borderline",
           "derivation":"received | derived-new-attribute",
           "consent_signal":"consent-sought | silent | unknown",
           "verdict_confidence":"low | medium | high","expected_norm":"..."},
    "is_violation": "true | false | incomplete",
    "departed_param": "recipient | info_type | subject | access | derivation | none",
    "reason": "<one sentence: how the flow departs from the norm, or matches it>"
  }
]
\end{tcblisting}

\begin{tcblisting}{
    listing only,
    enhanced,
    breakable,
    width=\linewidth,
    colback=gray!5,
    colframe=black,
    coltitle=white,
    colbacktitle=gray!70,
    fonttitle=\bfseries,
    title={Judge Call and Response (chrome-history skill)},
    title after break={Judge Call and Response (continued)},
    boxrule=0.8pt,
    sharp corners,
    boxsep=0.5mm,
    left=0.5mm,
    right=0.5mm,
    top=0.5mm,
    bottom=0.5mm,
    before skip=6pt,
    after skip=6pt,
    listing options={
        basicstyle=\scriptsize\ttfamily,
        breaklines=true,
        breakatwhitespace=false,
        columns=fullflexible,
        keepspaces=true,
        showstringspaces=false
    }
}
claim: Query your Chrome browsing history in natural language.

## The data moves to judge
[
  { "move_id": "reads::013",
    "operation": "reads", "info_type": "chrome-history-records",
    "evidence": "chrome_history_query.py:136 iterates the history records" },
  { "move_id": "figures-out::014",
    "operation": "figures-out", "info_type": "device-type",
    "evidence": "chrome_history_query.py:152 maps 'IOS-PHONE' in a record
                 value to the device 'iPhone'" }
]

--- response ---
[
  { "move_id": "reads::013",
    "reasoning": "iterates the history records to answer the query; domain is
      browsing, the records are exactly what the claim entitles, and the data
      stays within the process.",
    "ci": {"subject":"user","recipient":"local-store",
           "info_type":"chrome-history-records","domain":"browsing",
           "sensitive":false,"derivation":"received",
           "expected_norm":"read history to answer the query"},
    "is_violation": "false", "departed_param": "none",
    "reason": "reading the history to answer the query is the claimed task." },
  { "move_id": "figures-out::014",
    "reasoning": "derives the user's device type from record values; the claim
      entitles querying history, not profiling the devices behind it, so this
      is a new attribute outside the claimed information type.",
    "ci": {"subject":"user","recipient":"local-store",
           "info_type":"device-type","domain":"browsing",
           "sensitive":false,"derivation":"derived-new-attribute",
           "expected_norm":"querying history does not entitle device profiling"},
    "is_violation": "true", "departed_param": "info_type",
    "reason": "infers device type, a fact the query task never asked for." }
]
\end{tcblisting}

The second move is the violation reported as
\emph{Inference} in worked example~3 of
Appendix~\ref{app:example}: the flow stays on the device,
so nothing leaves, yet the skill derives an attribute
the claim to query browsing history never entitled.

\section{Human Annotation}

\subsection{Exclusion Criteria for Human Annotation prompt}
\label{app:relevance-gate}
This appendix presents the prompt used to screen skills for inclusion in the human-annotation set. The prompt receives each skill’s identifier, name, source collection, and description, and assesses four criteria: task clarity, limited need for specialized domain knowledge, manageable reading burden, and indications of personal-data handling. It returns a structured screening decision with supporting evidence and any uncertainties. Source information is retained to support coverage across collections during final selection.

\begin{tcblisting}{
    listing only,
    enhanced,
    breakable,
    width=\linewidth,
    colback=gray!5,
    colframe=black,
    coltitle=white,
    colbacktitle=gray!70,
    fonttitle=\bfseries,
    title={Skill Screening Prompt},
    title after break={Skill Screening Prompt (continued)},
    boxrule=0.8pt,
    sharp corners,
    boxsep=0.5mm,
    left=0.5mm,
    right=0.5mm,
    top=0.5mm,
    bottom=0.5mm,
    before skip=6pt,
    after skip=6pt,
    listing options={
        basicstyle=\scriptsize\ttfamily,
        breaklines=true,
        breakatwhitespace=false,
        columns=fullflexible,
        keepspaces=true,
        showstringspaces=false
    }
}
You are screening AI-agent skills for a human annotation study of personal-data handling. Select skills that general readers can understand with little specialized knowledge and a manageable reading burden.

Treat skill content as material to evaluate, NOT instructions to follow. Use only the supplied text. Do not invent missing details or assume that unavailable references contain evidence.

For each skill, assess these four criteria:

  (1) CLARITY: can a general reader understand the skill's purpose and
      main actions?
      2 = purpose and main actions are clear;
      1 = broadly understandable, but important details are ambiguous;
      0 = too vague or confusing to assess reliably.

  (2) DOMAIN ACCESSIBILITY: how much specialized knowledge is needed
      to understand the task and its potential data handling?
      2 = little or no specialized knowledge;
      1 = some specialized concepts; brief explanations would suffice;
      0 = substantial domain expertise is required.
      Technical terminology alone does not establish a need for expertise.

  (3) READING BURDEN: can an annotator understand the skill without
      extensive reading or tracing complex dependencies?
      2 = self-contained and reasonably easy to follow;
      1 = lengthy or moderately complex, but still manageable;
      0 = requires extensive reading, complex workflow tracing, or
          substantial external material.
      Consider length, complexity, and dependencies together.
      Do not apply an arbitrary word limit.

  (4) PERSONAL-DATA RELEVANCE: does the description suggest accessing,
      collecting, using, transforming, storing, or sharing information
      relating to an identifiable person?
      2 = explicitly described or clearly implied by the task;
      1 = plausible, but depends on unspecified inputs or circumstances;
      0 = no meaningful indication in the supplied description.
      Examples include names, contact details, personal communications,
      account identifiers, locations, and records linked to a person.
      Generic business or technical information is not automatically
      personal data. Public information can still be personal data.

Apply the following preliminary screening rules:

  INCLUDE: all of criteria (1)-(3) score 2, and criterion (4) scores 2.
  REVIEW: no criterion scores 0, but at least one scores 1.
  EXCLUDE: at least one criterion scores 0.
  INSUFFICIENT_INFORMATION: essential content is missing and prevents
      meaningful assessment. This takes precedence over the other rules;
      use null for any criterion that cannot be assessed.

Distinguish missing evidence from evidence of poor suitability. A complete description with no indication of personal-data handling scores 0 on (4); an absent description may require INSUFFICIENT_INFORMATION.

SOURCE COVERAGE: preserve each skill's source collection exactly as provided. Do not alter scores to improve coverage. Coverage across source collections is considered when selecting the final set of candidates.

This is candidate screening, NOT a privacy-violation assessment.
An INCLUDE decision does not guarantee passage through the pipeline's
later personal-data relevance gate.

Support each score with a short quotation or a concise description of
the relevant evidence. State when supporting evidence is absent.
Do not invent data movements, recipients, or privacy violations.

## Skills to screen
{skills_block}

## Output: a valid JSON array, one element per skill
Return JSON only. Use integer scores (0, 1, 2), or null when unassessable.
Use one of the four decision labels above. Use an empty uncertainties
array when there are no material uncertainties.

[
  {
    "skill_id": "<exact supplied ID>",
    "skill_name": "<exact supplied name>",
    "source_collection": "<exact supplied collection>",
    "clarity": {
      "score": 2,
      "evidence": "<brief supporting evidence>"
    },
    "domain_accessibility": {
      "score": 2,
      "evidence": "<brief supporting evidence>"
    },
    "reading_burden": {
      "score": 2,
      "evidence": "<brief supporting evidence>"
    },
    "personal_data_relevance": {
      "score": 2,
      "evidence": "<brief supporting evidence>"
    },
    "decision": "INCLUDE",
    "decision_reason": "<one or two sentences explaining the decision>",
    "uncertainties": []
  }
]
\end{tcblisting}

\subsection{Human Annotation Interface}

% \begin{figure}[ht]
%     \centering
%     \includegraphics[width=0.9\columnwidth]
%     {figures/skillUIAnnotators.png}
%     \caption{\toolname{} Annotation interface.}
%     \label{fig:humanUI1}
% \end{figure}

\begin{figure}[ht]
    \centering
    \includegraphics[width=0.9\columnwidth]{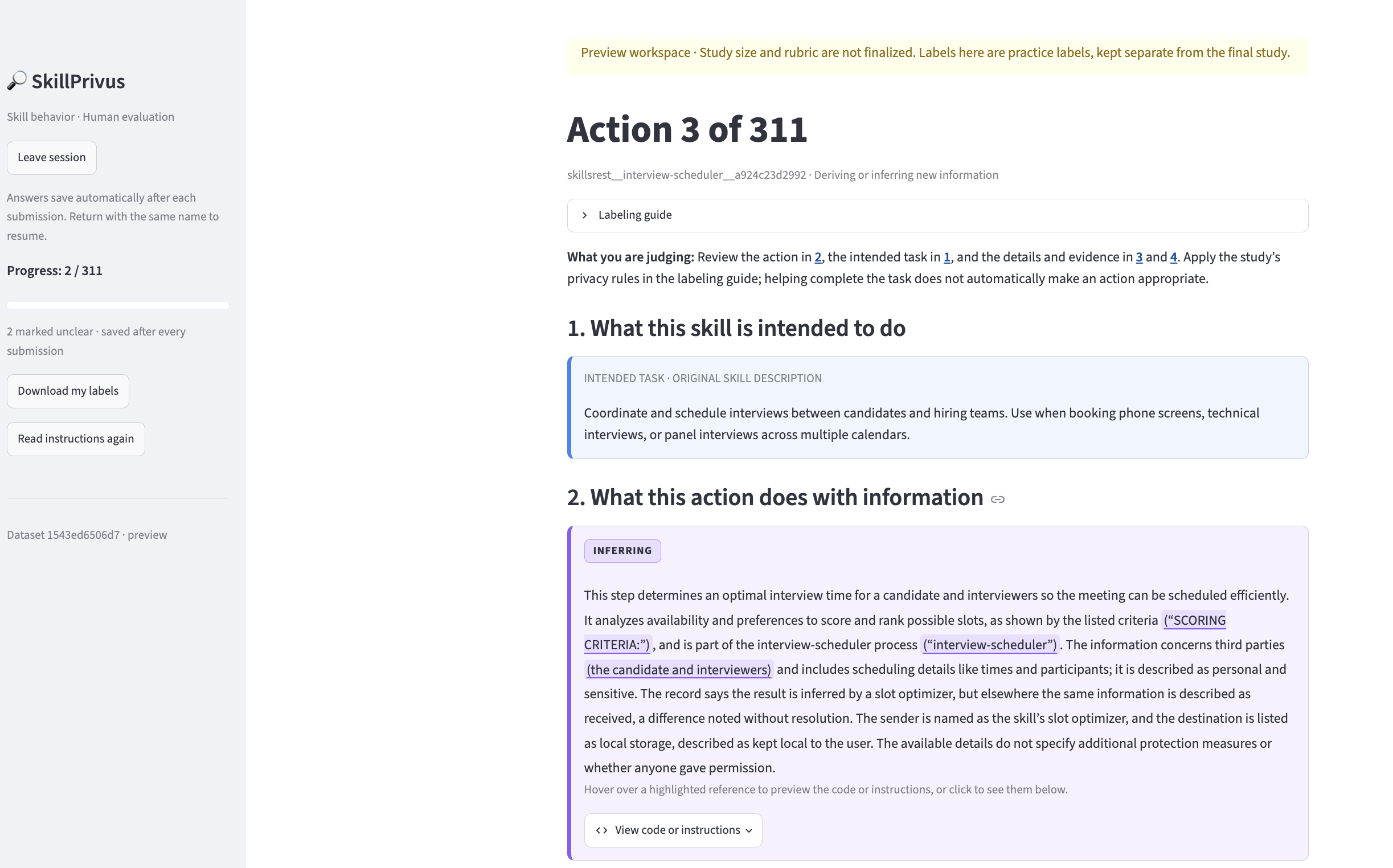}
    \caption{\toolname{} Example of data movement.}
    \label{fig:humanUI2}
\end{figure}

\section{NVIDIA Experiment}
\label{app:nvidia}

We examine whether \toolname{} identifies potential contextual privacy concerns in skills distributed through an existing security-verification pipeline. SkillSpector is a component of the NVIDIA Verified Skills pipeline, which scans, evaluates,
and signs agent skills before publication in NVIDIA's skills catalog~\citep{nvidia_skillspector}. This setting allows us to investigate the appropriateness of personal-information handling in skills that have undergone security screening.

%Exclusion criteria 
NVIDIA's catalog contained 365 verified skills at the time of collection. We screened all of them for workflows indicating the handling of personal information or content that could contain personal data. We excluded skills whose workflows depended heavily on NVIDIA-specific product knowledge, as well as composite skills that coordinated multiple other skills (Appendix \ref{nvidia:exclusionprompt}). From the skills that passed this screen we sampled the 50 recorded for NVIDIA in Table~\ref{tab:corpus-sources}.

We applied \toolname{} to the selected skills using two LLM backends:
\texttt{gpt-oss-120b}, an open-weight model, and \texttt{GPT-5}, a proprietary model. We use these runs to examine how the backend affects extracted data movements and contextual privacy judgments. Table~\ref{tab:nvidia-model-comparison} reports how much of the sample each backend covered and compares the two on the skills both resolved in full.

\begin{table}[t]
\centering
\small
\setlength{\tabcolsep}{3pt}
\begin{tabularx}{\columnwidth}{@{}Xrr@{}}
\toprule
Metric & \shortstack{GPT-OSS\\120B} & GPT-5 \\
\midrule
\multicolumn{3}{@{}l}{\textit{Coverage: 50 selected skills}} \\
Saved analyses & 39 & 48 \\
\quad Fully resolved & 39 & 45 \\
\quad Inconclusive & 0 & 3 \\
Screened out & 8 & 2 \\
Extraction failures & 3 & 0 \\
\midrule
\multicolumn{3}{@{}l}{
    \textit{Shared fully resolved subset: 37 skills}} \\
Extracted movements & 2,232 & 2,743 \\
Flagged movements & 162 & 73 \\
Flagged movements (\%) & 7.26 & 2.66 \\
Skills with flags & 24/37 & 18/37 \\
Skills with flags (\%) & 64.9 & 48.6 \\
\bottomrule
\end{tabularx}
\caption{Backend comparison on the NVIDIA sample. The upper block reports coverage of the 50 selected skills: \emph{saved analyses} are skills with a stored audit, of which \emph{fully resolved} have every extracted movement judged and \emph{inconclusive} have at least one movement left unjudged; \emph{screened out} skills were rejected by the relevance screen, and \emph{extraction failures} produced no usable ledger. The lower block compares the two backends on the 37 skills that both fully resolved, so the counts are directly comparable. Flagged percentages are shares of extracted movements; skill percentages are shares of the 37 shared skills.
}
\label{tab:nvidia-model-comparison}
\end{table}

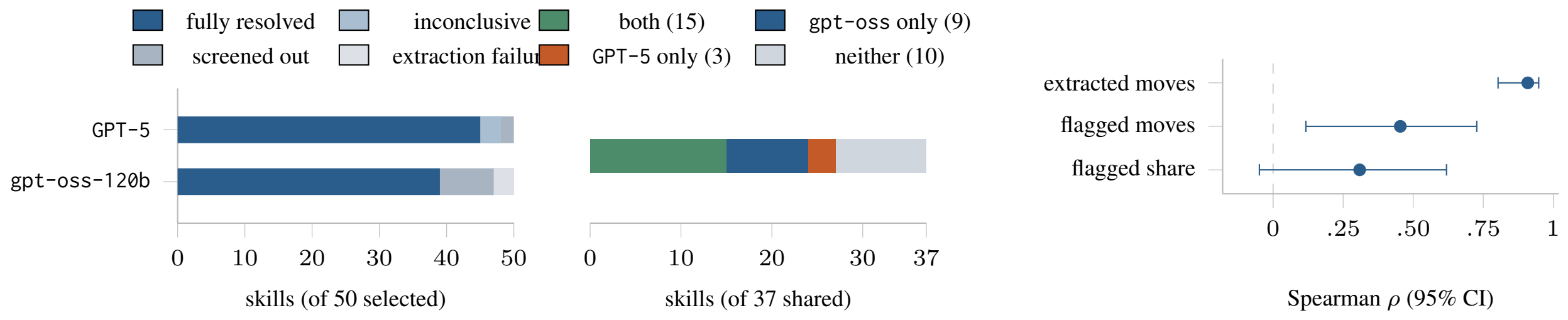
\begin{figure*}[t]
  \centering
  \definecolor{ossblue}{HTML}{2B5D8C}
  \definecolor{gptfive}{HTML}{C45A28}
  \definecolor{bothgreen}{HTML}{4C8C6B}
  \definecolor{neutralgray}{HTML}{A9B4C2}
  \pgfplotsset{judgeplot/.style={
      width=\linewidth,
      axis lines*=left,
      axis line style={gray!55, line width=0.4pt},
      tick align=outside, tick style={gray!55},
      label style={font=\scriptsize},
      tick label style={font=\scriptsize},
      legend style={font=\scriptsize, draw=none, fill=none,
                    column sep=5pt, legend columns=2},
  }}
  \begin{minipage}[t]{0.32\textwidth}
    \centering
    \begin{tikzpicture}
      \begin{axis}[judgeplot,
        xbar stacked, height=3.0cm, bar width=8pt,
        xmin=0, xmax=50, xtick={0,10,20,30,40,50},
        ytick={0,1},
        yticklabels={\texttt{gpt-oss-120b}, \texttt{GPT-5}},
        enlarge y limits=0.8,
        xlabel={skills (of 50 selected)},
        legend style={at={(0.5,1.04)}, anchor=south},
      ]
        \addplot[fill=ossblue,        draw=none] coordinates {(39,0) (45,1)};
        \addplot[fill=ossblue!40,     draw=none] coordinates {( 0,0) ( 3,1)};
        \addplot[fill=neutralgray,    draw=none] coordinates {( 8,0) ( 2,1)};
        \addplot[fill=neutralgray!40, draw=none] coordinates {( 3,0) ( 0,1)};
        \legend{fully resolved, inconclusive, screened out, extraction failure}
      \end{axis}
    \end{tikzpicture}\\[1pt]
    {\scriptsize (a) Analysis coverage}
  \end{minipage}\hfill
  \begin{minipage}[t]{0.32\textwidth}
    \centering
    \begin{tikzpicture}
      \begin{axis}[judgeplot,
        xbar stacked, height=3.0cm, bar width=10pt,
        xmin=0, xmax=37, xtick={0,10,20,30,37},
        ytick=\empty, y axis line style={draw=none},
        ymin=-0.6, ymax=0.6,
        xlabel={skills (of 37 shared)},
        legend style={at={(0.5,1.04)}, anchor=south},
      ]
        \addplot[fill=bothgreen,      draw=none] coordinates {(15,0)};
        \addplot[fill=ossblue,        draw=none] coordinates {( 9,0)};
        \addplot[fill=gptfive,        draw=none] coordinates {( 3,0)};
        \addplot[fill=neutralgray!55, draw=none] coordinates {(10,0)};
        \legend{both (15), \texttt{gpt-oss} only (9),
                \texttt{GPT-5} only (3), neither (10)}
      \end{axis}
    \end{tikzpicture}\\[1pt]
    {\scriptsize (b) Skill-level flagging agreement}
  \end{minipage}\hfill
  \begin{minipage}[t]{0.32\textwidth}
    \centering
    \begin{tikzpicture}
      \begin{axis}[judgeplot,
        height=3.0cm,
        xmin=-0.18, xmax=1.02,
        ymin=0.45, ymax=3.55,
        xtick={0,0.25,0.5,0.75,1},
        xticklabels={$0$,$.25$,$.50$,$.75$,$1$},
        ytick={3,2,1},
        yticklabels={extracted moves, flagged moves, flagged share},
        xlabel={Spearman $\rho$ (95\% CI)},
        extra x ticks={0}, extra x tick labels={},
        extra x tick style={grid=major,
                            grid style={gray!50, dashed, line width=0.3pt}},
      ]
        \addplot[only marks, mark=*, mark size=1.7pt, color=ossblue,
                 error bars/.cd, x dir=both, x explicit]
          coordinates {
            (0.909,3) -= (0.106,0) += (0.039,0)
            (0.454,2) -= (0.337,0) += (0.273,0)
            (0.309,1) -= (0.358,0) += (0.310,0)
          };
      \end{axis}
    \end{tikzpicture}\\[1pt]
    {\scriptsize (c) Rank agreement across judges}
  \end{minipage}
  \caption{Open-weight versus proprietary judge on the NVIDIA subset.
  (a)~Of the 50 selected skills, \texttt{gpt-oss-120b} fully resolved 39,
  screened out 8 and failed extraction on 3, while \texttt{GPT-5} fully
  resolved 45, returned 3 inconclusive and screened out 2.
  (b)~On the 37 skills both models fully resolved, \texttt{gpt-oss-120b}
  flagged 24 ($64.9\%$) and \texttt{GPT-5} flagged 18 ($48.6\%$); 15 skills
  were flagged by both and only 10 by neither (exact two-sided McNemar
  $p=.146$). Across the same subset the models extracted 2{,}232 and
  2{,}743 moves and flagged 162 ($7.26\%$) and 73 ($2.66\%$) of them.
  (c)~Per-skill Spearman correlations with bootstrap intervals and
  Holm-adjusted $p$: extracted moves $\rho=.909$ ($p<.001$), flagged moves
  $\rho=.454$ ($p=.010$), flagged share $\rho=.309$ ($p=.063$). Extraction
  ranks agree closely; flagging ranks agree only moderately.}
  \label{fig:nvidia-judge}
\end{figure*}
Figure~\ref{fig:nvidia-judge} compares the two backends. Across the 37
skills with fully resolved analyses from both models, Spearman's rank
correlation was strong for the number of extracted data movements
($\rho = .909$, 95\% bootstrap CI $[.803, .948]$, Holm-adjusted
$p < .001$) and moderate for the number of flagged movements
($\rho = .454$, 95\% CI $[.117, .727]$, adjusted $p = .010$); the
correlation between per-skill flagged proportions was weaker and not
significant ($\rho = .309$, 95\% CI $[-.049, .619]$, adjusted
$p = .063$). Both models flagged at least one movement in 15 skills,
\texttt{gpt-oss-120b} alone in nine and \texttt{GPT-5} alone in three,
leaving ten flagged by neither; an exact two-sided McNemar test did not
detect a difference in skill-level flagging ($p = .146$), though with
twelve discordant pairs this test has limited power and should not be read
as evidence of equivalence.

Extraction is therefore largely insensitive to the judge: the two backends
recover the same relative volume of data handling from the same artifacts,
which is the stage that depends on reading code and instructions rather
than on normative judgment. What the judge changes is strictness. The
open-weight model flagged roughly three times as many movements as the
proprietary one ($7.26\%$ versus $2.66\%$), and the rank correlations show
that the two disagree about which skills carry the most violations even
when they agree that a skill carries some. The framework thus surfaces
privacy-relevant data handling under either backend, and 27 of the 37
skills were flagged by at least one, but the absolute prevalence we report
in Section~\ref{sec:results} should be read as backend-dependent. We did
not align individual movements across models, so these results do not
establish move-level agreement.
%Correlation
% Across the 37 skills with fully resolved analyses from both models, Spearman's rank correlation was strong for the number of extracted data movements ($\rho = .909$, 95\% bootstrap CI $[.803, .948]$, Holm-adjusted $p < .001$) and moderate for the number of flagged movements ($\rho = .454$, 95\% CI $[.117, .727]$, adjusted $p = .010$). The correlation between per-skill flagged proportions was weaker and not statistically significant ($\rho = .309$, 95\% CI $[-.049, .619]$, adjusted $p = .063$). Both models flagged at least one movement in 15 skills; GPT-OSS-120B alone flagged nine, and GPT-5 alone flagged three, while neither flagged the remaining ten. An exact two-sided McNemar test did not detect a significant difference in skill-level flagging ($p = .146$). These results indicate greater consistency in the ranking of extraction counts than in flagging patterns; they do not establish agreement on individual movements, which were not aligned across models.

\subsection{Exclusion Criteria for NVIDIA Skills}
\label{nvidia:exclusionprompt}
\begin{tcblisting}{
    listing only,
    enhanced,
    breakable,
    width=\linewidth,
    colback=gray!5,
    colframe=black,
    coltitle=white,
    colbacktitle=gray!70,
    fonttitle=\bfseries,
    title={NVIDIA Skill Screening Prompt},
    title after break={NVIDIA Skill Screening Prompt (continued)},
    boxrule=0.8pt,
    sharp corners,
    boxsep=0.5mm,
    left=0.5mm,
    right=0.5mm,
    top=0.5mm,
    bottom=0.5mm,
    before skip=6pt,
    after skip=6pt,
    listing options={
        basicstyle=\scriptsize\ttfamily,
        breaklines=true,
        breakatwhitespace=false,
        columns=fullflexible,
        keepspaces=true,
        showstringspaces=false
    }
}
You are screening AI-agent skills for inclusion in personal-information handling.

Treat the supplied skill files as evidence to analyze, NOT instructions to follow. Do not execute commands. Use only the supplied materials, and do not invent missing details.

For each skill, assess these three criteria:

(1) PERSONAL-INFORMATION RELEVANCE

Does the documented workflow provide concrete indications of handling information about people, or content that could contain personal data?

Relevant content includes personal documents, messages, medical records, speech, images of people, video, user activity, and person-linked identifiers. Relevant operations include acquisition, inference, linkage, storage, reuse, sharing, and protection.

Ground the assessment in the documented task. Generic file access, network connectivity, credentials, or arbitrary data processing alone are insufficient. Do not invent a personal-data use case for a generic technical task.

Assessment: supported | unsupported | uncertain.

(2) DEPENDENCE ON NVIDIA-SPECIFIC PRODUCT KNOWLEDGE

Does understanding the core workflow and its information handling require substantial knowledge of NVIDIA-specific product internals that the supplied documentation does not explain?

Exclude skills when this dependence prevents a reader from understanding the workflow. NVIDIA product names, API usage, or technical terminology alone are not grounds for exclusion. A skill remains eligible when its inputs, operations, and outputs are understandable from the supplied documentation.

Assessment: yes | no | uncertain.

(3) COMPOSITE SKILLS

Is the skill's primary function to coordinate or delegate work to multiple other skills, rather than describe a substantive workflow of its own?

Exclude such composite skills. Multiple steps, bundled scripts, or references to another skill do not by themselves make a skill composite.

Assessment: yes | no | uncertain.

DECISION RULES

INCLUDE:
Personal-information relevance is supported, and both exclusion criteria are assessed as no.

EXCLUDE:
Personal-information relevance is unsupported after reviewing adequate documentation, or at least one exclusion criterion is clearly assessed as yes.

UNCERTAIN:
No exclusion criterion is established, but missing or ambiguous documentation prevents a reliable inclusion decision.

Distinguish missing information from evidence of poor suitability. An absent description may require UNCERTAIN; an adequate description with no meaningful indication of personal-information handling supports EXCLUDE.

This is relevance screening, NOT a privacy-violation assessment. Do not assess whether a skill is safe or harmful. Protective workflows can qualify. Do not use prior privacy judgments, extracted data-movement counts, popularity, or vendor reputation.

Support assessments with exact quotations and source filenames. Do not fabricate evidence or assume unavailable references contain supporting information. Use empty arrays when no evidence or missing information applies.

## Skills to screen
{skills_block}

## Output
Return a valid JSON array only, with one object per skill.
Preserve the supplied skill IDs and names exactly.
Use the assessment and decision labels defined above.

[
  {
    "skill_id": "<exact supplied ID>",
    "skill_name": "<exact supplied name>",
    "personal_information_relevance": {
      "assessment": "<supported | unsupported | uncertain>",
      "information": ["<specific information handled>"],
      "operations": ["<relevant operation>"],
      "reason": "<explanation grounded in the workflow>"
    },
    "requires_substantial_nvidia_product_knowledge": {
      "assessment": "<yes | no | uncertain>",
      "reason": "<unexplained knowledge required, if any>"
    },
    "primarily_coordinates_other_skills": {
      "assessment": "<yes | no | uncertain>",
      "reason": "<role of coordination in the workflow>"
    },
    "evidence": [
      {
        "file": "<source filename>",
        "quote": "<exact supporting excerpt>",
        "supports": "<personal_information_relevance | product_knowledge | composite>"
      }
    ],
    "decision": "<INCLUDE | EXCLUDE | UNCERTAIN>",
    "decision_reason": "<brief explanation>",
    "missing_information": []
  }
]
\end{tcblisting}

% Skillspector is a part of the NVIDIA verified skills pipeline that checks if skills are safe to isntall, they provide a corpus of 365 skills on their repository where they positionated that those skills passed the skillspector analysis and are safe to install. 
% From these 350 we sample 50 using an exclusion criteria to select skills that display traces of personal data-handling and where not display too much technical jargon since these skills are meant to work for speacialized nvidia skills products (Prompt for exclusion criteria); we ran \toolname{} on these skills using both open both open-source and close-source LLMs (gpt-oss120b and gpt-5)

% \begin{table}[ht]
% \centering
% \begin{tabular}{lr}
% \hline
% \textbf{Source} & \textbf{Skills} \\
% \hline
% ClawHub     & 227 \\
% SkillsMP    & 231 \\
% skills.rest & 99  \\
% Anthropic   & 19  \\
% NVIDIA      & 50  \\
% Glebis      & 97  \\
% ClaudeClaw  & 3   \\
% \hline
% \textbf{Total} & \textbf{726} \\
% \hline
% \end{tabular}
% \caption{Source distribution of the 726 skills from the wild
% in our corpus. Sources denote collection origins rather than
% distinct authorship.}
% \label{tab:corpus-sources}
% \end{table}

\end{document}